\documentclass[11pt]{article}

\usepackage[margin=1in]{geometry}
\usepackage{amsmath,amssymb}
\usepackage{graphicx}
\usepackage{booktabs}
\usepackage{multirow}
\usepackage{hyperref}
\usepackage{caption}
\usepackage{subcaption}
\usepackage{cite}
\graphicspath{{zigzag/main/publication/}}

\title{Topological Signatures of Context-Level Reliability in TabPFN}
\author{
\textbf{James Hu, Mahdi Ghelichi} \\
Model Development Innovation, Risk Management \\
TD Bank, Toronto, Canada \\
\texttt{\{james.hu, mahdi.ghelichi\}@td.com}
}
\date{}

\begin{document}
\maketitle

\begin{abstract}
TabPFN is a transformer-based foundation model for tabular prediction that performs inference without task-specific training by conditioning on a support set and query inputs. Despite its strong empirical performance, its internal behavior on structurally difficult tabular geometries remains poorly understood. We study this behavior using zigzag persistent homology, treating TabPFN layer representations as evolving point clouds. We construct a controlled benchmark of synthetic tabular tasks with known true probabilities and varied intrinsic topology, including warped circles, tori, spheres, Hopf links, trefoil knots, and Swiss rolls. Across these tasks, we find that the topology of TabPFN's internal representation geometry is strongly associated with dataset-level reliability; for example, the zeroth homology group $H_0$ fragmentation count correlates positively with mean absolute residual across controlled tasks, and this association strengthens in a high-resolution warped circle case study at large sample size. Harder geometries induce a dual topological signature: increased $H_1$ loop activity and increased $H_0$ fragmentation, while the $H_1$ persistence becomes shorter-lived. These descriptors correlate with Bayes error, mean absolute residuals, and overconfidence. Our results suggest that zigzag persistence diagnoses the reliability of the inferred in-context task geometry and provides a context-level view of when TabPFN operates in topologically stressed regimes.
\end{abstract}

\section{Introduction}
\label{sec:introduction}
Foundation models for tabular data have recently become a promising alternative to task-specific model training. TabPFN, in particular, performs tabular prediction by conditioning on a set of labeled context rows and unlabeled query rows, using a transformer architecture trained on large collections of synthetic tabular tasks \cite{hollmann2023tabpfn, hollmann2025tabpfnv2, tabpfn_v2_5, tabpfn_v3}. This makes TabPFN an in-context learner for tabular data: at inference time, the model does not update its weights, but instead infers a task from the provided support set.

Recent studies have begun to examine TabPFN's robustness under controlled tabular perturbations \cite{ye2025closerlooktabpfnv2, hu2026noiseimmunityincontexttabular}, but its inference mechanism raises a different interpretability question. When TabPFN works, what kind of internal task geometry does it construct? When TabPFN becomes unreliable, how does the geometry of the entire in-context task change inside the transformer?

Additional work has been studying the combination of topological data analysis (TDA) with TabPFN in downstream prediction settings. For example, persistent-homology features extracted from medical imaging data have been successfully used as inputs to TabPFN models, with loop and cavity based topological descriptors contributing substantially to predictive performance \cite{han2026topological}. These results suggest that topology can provide meaningful information for tabular foundation models, but leave open the complementary question we study here: whether the topology of TabPFN's own internal representations is itself informative.

Recent work on large language models has shown that the geometry of hidden representations can reveal how transformers reorganize inputs across layers \cite{gardinazzi2025persistent}. In particular, zigzag persistence has been proposed as a way to track topological features as point clouds evolve through model depth, rather than analyzing each layer independently \cite{carlsson2010zigzag}. A related line of work, HalluZig, models layerwise LLM attention matrices as evolving graphs and uses zigzag persistence to study reliability-relevant internal dynamics \cite{samaga2026halluzig}. These studies suggest that topological summaries of internal transformer dynamics can expose structure that is not visible from output probabilities alone.

In this paper, we extend the representation-topology perspective to TabPFN. Unlike standard language prompts, TabPFN inputs define an in-context tabular task: support rows, support labels, and query rows jointly determine the prediction problem. We therefore ask whether the topology of TabPFN's internal representations reflects the reliability of this inferred task geometry. We focus on the first two homology dimensions. The zeroth homology group, $H_0$, captures connected components; in TabPFN's layerwise representation graph, it measures how the in-context task separates into or merges across coarse groups. The first homology group, $H_1$, captures one-dimensional cycles or loops; here it measures cyclic or entangled structure in the representation graph as TabPFN processes more difficult tabular tasks.

In this paper, we make the following contributions:
\begin{enumerate}
    \item We introduce a zigzag-persistence framework for analyzing TabPFN hidden representations across layers.
    \item We construct a controlled topology benchmark for tabular prediction with known true probabilities and varied intrinsic topology.
    \item We show that TabPFN representation topology predicts dataset-level reliability metrics, including observed error, Bayes error, mean absolute residuals, and overconfidence.
    \item We identify an $H_0/H_1$ ``scissors'' pattern---named for two topological quantities that move in opposite directions as difficulty rises: $H_1$ loop activity increases while the durable persistence of $H_0$ connected components decreases (with the $H_0$ event count rising alongside).
    \item We connect these findings to prior topological analyses of transformer representations and interpret TabPFN topology as the topology of an inferred in-context task geometry.
\end{enumerate}
The organization of the paper is as follows. Section 2 discusses related work on topological analysis and zigzag persistence  and their applications on transformers. In Section 3, we describe the experiment setups including zigzag construction and metrics definitions. Section 4 describes the synthetic datasets with different topology families used for the experiments. In Section 5, we present the experiment results. Further implications from the results, limitations, and future work are discussed in Section 6. Section 7 concludes the paper.

\section{Related Work}

\subsection{Tabular foundation models and in-context learning}
Classical tabular prediction has long been dominated by tree-based ensembles such as random forests and gradient-boosted trees \cite{randomforest,xgboost,catboost}, which remain strong baselines and often match or exceed early tabular deep-learning models \cite{grinsztajn2022whytree,gorishniy2021revisiting,tabnet}. TabPFN reframed the problem as in-context learning: a transformer \cite{attention} pretrained on large collections of synthetic tasks predicts on a new dataset in a single forward pass, without per-dataset training \cite{hollmann2023tabpfn,hollmann2025tabpfnv2}. This mechanism connects to a broader literature on how transformers learn in context \cite{ICL,von2023transformers}. A fast-growing line of work extends the paradigm by scaling it to larger and real-world tables \cite{tabdpt,tabpfn_v2_5,tabpfn_v3,orion_msp}, benchmarking it \cite{tabarena}, probing what it internalizes \cite{ye2025closerlooktabpfnv2,swelam2025does}, characterizing its robustness \cite{labelnoise_tabpfn,hu2026noiseimmunityincontexttabular}, and adapting it through fine-tuning \cite{tanna2026exploring}. The prospect of a single pretrained model that transfers across tables has made tabular foundation models a research priority \cite{van2024tabular}, with direct relevance to high-stakes tabular domains such as biomedical prediction \cite{kursa2014robustness} and financial risk management \cite{haeri2025riskrag}. Our contribution is complementary to this line: rather than improving or benchmarking TabPFN, we ask what its internal representation geometry reveals about \emph{when} its predictions can be trusted.

\subsection{Topology of neural representations}
Topological Data Analysis (TDA) provides tools for studying the shape of high-dimensional point clouds. Persistent homology tracks the birth and death of topological features across a filtration, producing barcodes or persistence diagrams that record connected components, loops, voids, and higher-dimensional structures \cite{edelsbrunner2002topological,zomorodian2005computing,carlsson2009topology,edelsbrunner2010computational}. Prior works such as \cite{rieck2019neural,naitzat2020topology} on neural-network topology has used Betti numbers, persistence diagrams, and topological descriptors to study how internal representations change through depth of the networks. These studies motivate our use of $H_0$ and $H_1$ as interpretable summaries of TabPFN representation topology.

\subsection{Zigzag persistence for transformer representations}
The closest methodological precedent is the recent zigzag persistence framework for large language models \cite{gardinazzi2025persistent}. That work treats last-token representations from many prompts as point clouds that evolve through the layers of decoder-only transformers. For each layer, the authors construct a k-nearest-neighbor graph (with default setup $k=4$), expand it into a simplicial complex, insert intersection complexes between adjacent layers, and compute zigzag persistence. Their descriptors reveal layerwise processing phases and support a layer-pruning criterion.

Our work adopts the same dynamic view, but applies it to a different object. In TabPFN, the evolving point cloud represents an in-context tabular task built from context and query rows. This changes the interpretation of the topology: the descriptors are not only summaries of hidden-state shape, but diagnostics of the reliability of the inferred task geometry.

\subsection{Topology and reliability in transformer models}
HalluZig demonstrates that zigzag persistence can be applied to transformer attention graphs for reliability analysis \cite{samaga2026halluzig}. The method converts layerwise attention matrices into graphs, applies zigzag persistence, and uses the resulting signature to characterize internal attention dynamics. Our work shares the broad reliability motivation, but studies hidden representation topology in TabPFN rather than attention topology in language models. The common theme is that evolving transformer-internal graphs or point clouds can reveal reliability-relevant structure.

\section{Experimental Method}

\subsection{Extracting TabPFN layer representations}
TabPFN processes a tabular task by conditioning on labeled context rows and unlabeled query rows. We extract hidden representations from each of the 12 transformer layers of TabPFN v2 \cite{hollmann2025tabpfnv2}. For each layer, the selected representations form a point cloud in hidden space. Depending on the experiment, the point cloud may consist of query representations, support representations, or other embedding scopes.

Our main experiments use query label-token representations as the primary view of the internal task geometry, with additional robustness checks across support feature-token and support label-token representations.

\subsection{Constructing a zigzag filtration}
For each layer $\ell$, we construct a k-nearest-neighbor graph from the representation point cloud and expand this graph into a simplicial complex $K_\ell$ by filling cliques up to a fixed maximum simplex dimension. Following prior zigzag representation work \cite{gardinazzi2025persistent}, we set $k=4$ as our default value.
Consecutive layer complexes are not nested, because nearest-neighbor relations can appear and disappear as representations move through the transformer. As used in  \cite{le2022persistent}, we therefore insert intersection complexes between adjacent layers:
\begin{equation}
K_0 \leftarrow K_0 \cap K_1 \rightarrow K_1
\leftarrow K_1 \cap K_2 \rightarrow \cdots \leftarrow K_{L-2}\cap K_{L-1}\rightarrow K_{L-1}.
\end{equation}
For TabPFN v2 \cite{hollmann2025tabpfnv2} with $L=12$ transformer layers, this produces 23 zigzag positions: 12 model-layer complexes and 11 intermediate transition complexes.

The output of zigzag persistence is a multiset of intervals $[b, d]$, where each interval represents a topological feature born at position $b$ and dying at position $d$. Following prior work on LLM zigzag persistence, we map raw zigzag intervals to effective model-layer intervals to avoid artifacts induced by alternating model and intermediate intersection layers \cite{carlsson2010zigzag,gardinazzi2025persistent}.

\subsection{Topological descriptors}
We compute descriptors in homology dimensions $H_0$ and $H_1$. Here, $H_0$ reflects coarse grouping, fragmentation, and connectedness of the representation graph. $H_1$ reflects higher-order relational complexity or entanglement in the representation topology.

Intuitively, $H_0$ and $H_1$ summarize two different ways TabPFN can arrange the in-context examples in its hidden space. $H_0$ counts connected components, so it tracks how the support and query rows group together: when the task is easy and aligned with the model's prior, the rows coalesce into a few clean, long-lived clusters, which is the geometry of a confidently separable decision problem. As difficulty rises, those clusters break into many small, short-lived fragments---the model can no longer hold the examples in a single coherent grouping, so the $H_0$ event count grows while the lifetime of each component shrinks. $H_1$ counts one-dimensional loops, which appear when the model wraps the examples back on themselves instead of laying them out along a simple separating direction; a loop is the signature of an entangled, non-linearly-separable arrangement, and more or longer-lived loops indicate a more convoluted internal decision surface. Read together, $H_0$ reports whether the model still perceives clean groups and $H_1$ reports whether it must bend the representation into cycles to accommodate the labels, so their joint movement (Section~\ref{sec:scissors}) describes how the inferred task geometry deforms as the input becomes harder.

We organize descriptors into three families:
\paragraph{Activity metrics.}
These measure how much topology is generated across the layer trajectory:
\begin{equation}
H_1 \, \text{area} = \sum_{[b,d]\in H_1}(d-b),
\end{equation}
with analogous total-persistence summaries for $H_0$. We also compute $H_0$ and $H_1$ bar counts and Betti-curve summaries.
\paragraph{Stability metrics.}
These compute how long topological features survive, including mean persistence, maximum persistence, and durable-feature fractions.
\paragraph{Timing metrics.}
These evaluate where features appear or resolve across model depth. We use early/middle/late persistence mass, birth-persistence heatmaps, and a persistence-weighted histogram $\pi_{\mathrm{hist}}(\alpha)$ analogous to the births' relative frequency descriptor used in prior LLM zigzag work \cite{gardinazzi2025persistent}. Unless otherwise noted, we report $\alpha=1$, which weights features by their effective persistence.

\subsection{True-probability reliability metrics}
One limitation of using real datasets for our experiments is that the dataset's true conditional probability $P^*(y\mid x)$ is unknown. To study calibration and overconfidence precisely, we construct synthetic datasets for classification problems with known true probabilities. For each test row $x$, we compute
\begin{equation}
\hat p_{\mathrm{pred}} = \hat p(\hat y\mid x),
\end{equation}
where $\hat y$ is TabPFN's predicted class. We also compute the true probability of that predicted class:
\begin{equation}
p^*_{\mathrm{pred}} = P^*(\hat y\mid x).
\end{equation}
The signed residual is
\begin{equation}
r(x)=\hat p_{\mathrm{pred}}-p^*_{\mathrm{pred}}.
\end{equation}
Positive residuals indicate overconfidence relative to the truth. The main calibration quantity used in the results is the dataset-level mean absolute residual (MAR),
\begin{equation}
\mathrm{MAR}=\frac{1}{n}\sum_{i=1}^n |r(x_i)|,
\end{equation}
which measures the typical gap between TabPFN's confidence and the true probability of the predicted class. We also report metrics such as observed error, Bayes error, overconfidence rate, and wrong-overconfidence rate.

Observed error is the fraction of test points for which TabPFN's predicted label differs from the sampled test label. We use Bayes error as shorthand for Bayes-label error: the fraction of test points for which TabPFN's predicted label differs from the Bayes-optimal label $\mathbf{1}[P^*(y{=}1\mid x)\ge 1/2]$. Thus Bayes error measures disagreement with the true decision rule, not the irreducible Bayes risk of the data-generating process. The overconfidence rate is the fraction of test points with $r(x)>0.10$, and the wrong-overconfidence rate further requires the prediction to be incorrect.

\paragraph{Statistical analysis.}
Unless otherwise noted, we quantify monotonic associations between a topological descriptor and a reliability metric with the Spearman rank-correlation coefficient $\rho$, which ranges from $-1$ for a perfectly decreasing monotone relationship to $+1$ for a perfectly increasing one. Hard-versus-easy contrasts are reported as Cohen's $d$, a standardized mean difference. Significance is assessed with the Mann--Whitney $U$ test for group comparisons and the Wilcoxon signed-rank test for paired base-versus-fine-tuned comparisons. We use the conventional thresholds ${}^{*}p<0.05$, ${}^{**}p<0.01$, and ${}^{***}p<0.001$, and write ``n.s.''\ for $p\ge0.05$.

\section{Controlled Topology Benchmark}

We construct six synthetic topology families designed to stress TabPFN across different geometric and topological regimes.

\subsection{Topology families}
\begin{figure}[t]
\centering
\begin{subfigure}{0.31\textwidth}\includegraphics[width=\linewidth]{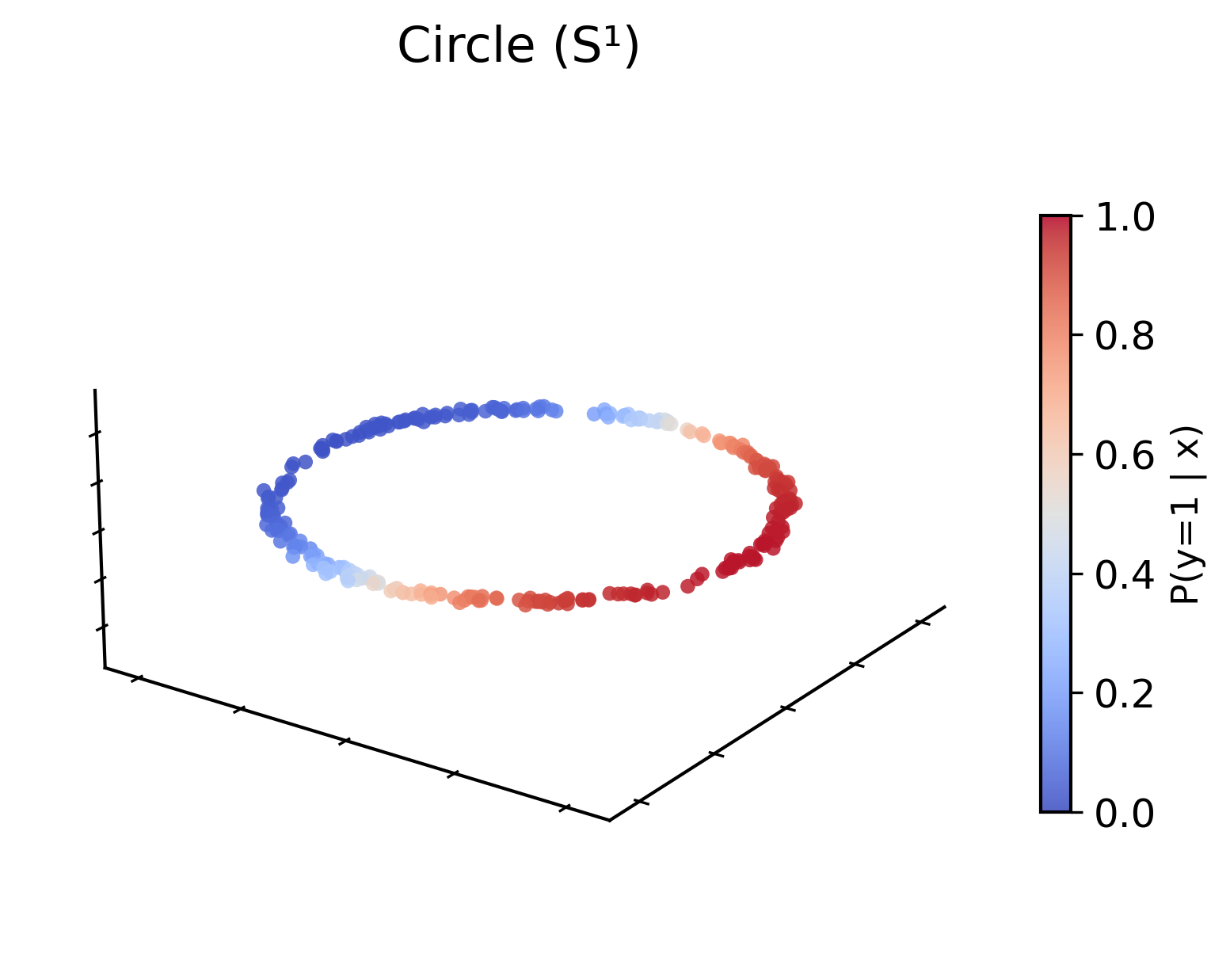}\caption{Warped circle ($S^1$)}\end{subfigure}\hfill
\begin{subfigure}{0.31\textwidth}\includegraphics[width=\linewidth]{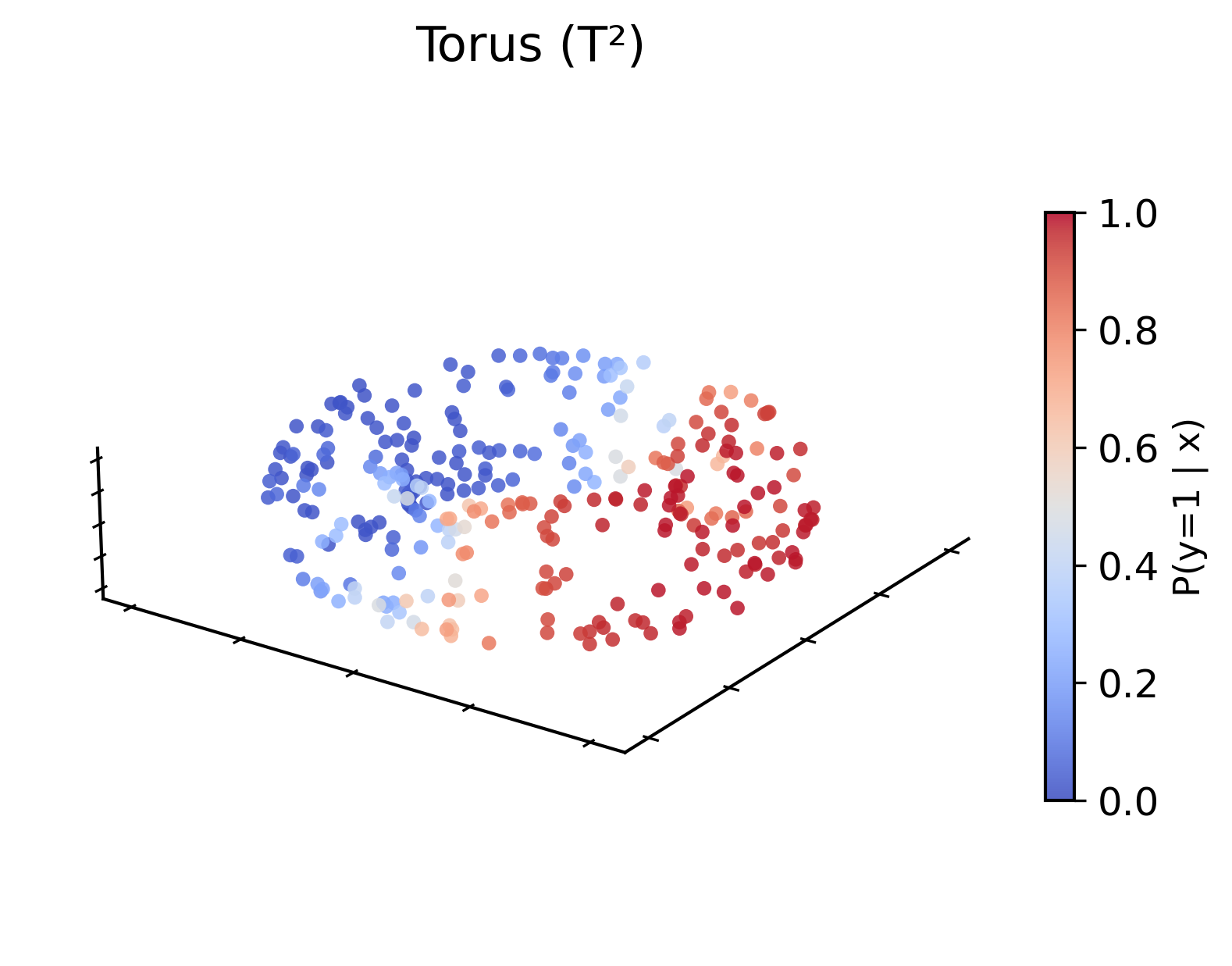}\caption{Torus ($T^2$)}\end{subfigure}\hfill
\begin{subfigure}{0.31\textwidth}\includegraphics[width=\linewidth]{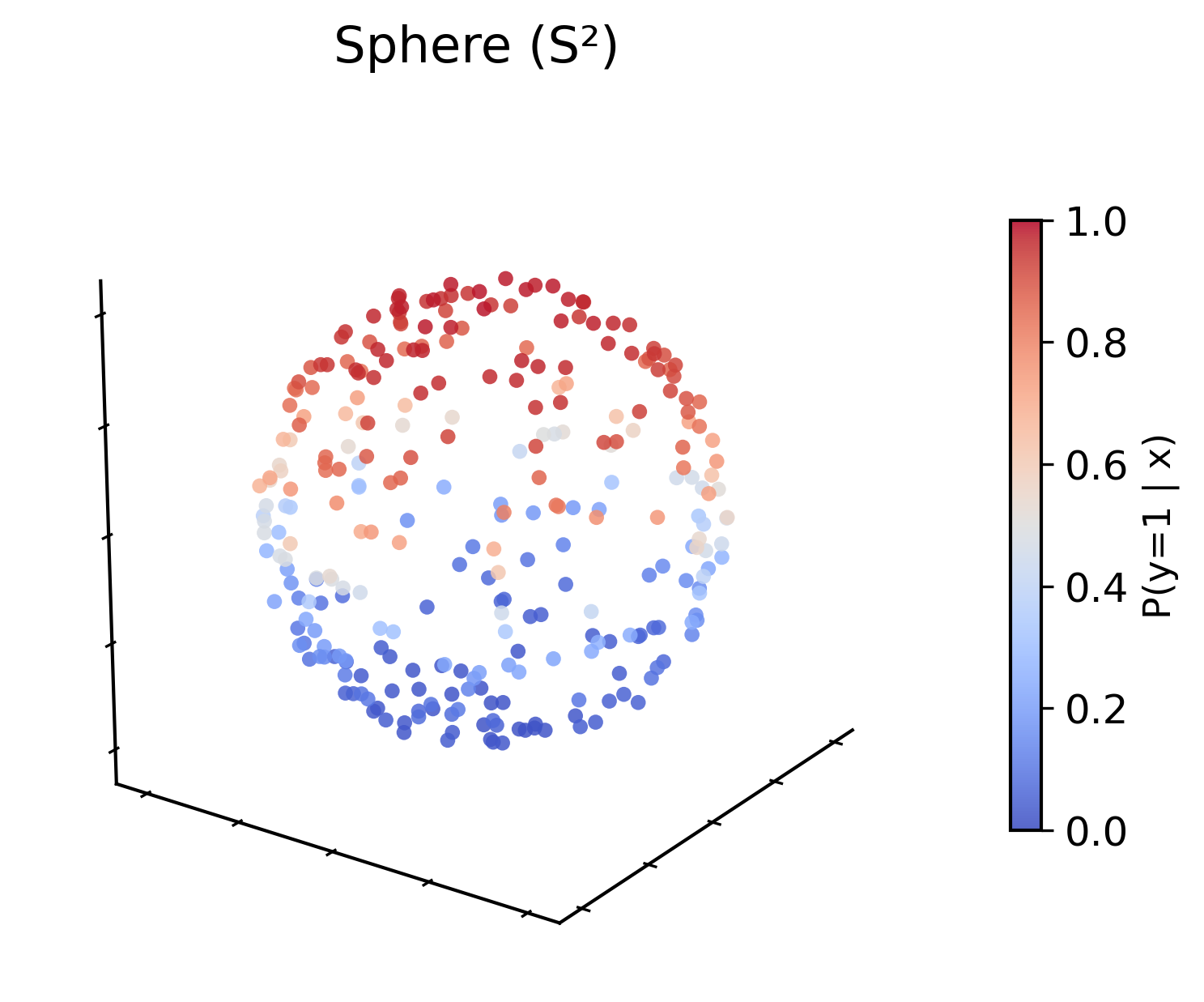}\caption{Sphere ($S^2$)}\end{subfigure}

\medskip
\begin{subfigure}{0.31\textwidth}\includegraphics[width=\linewidth]{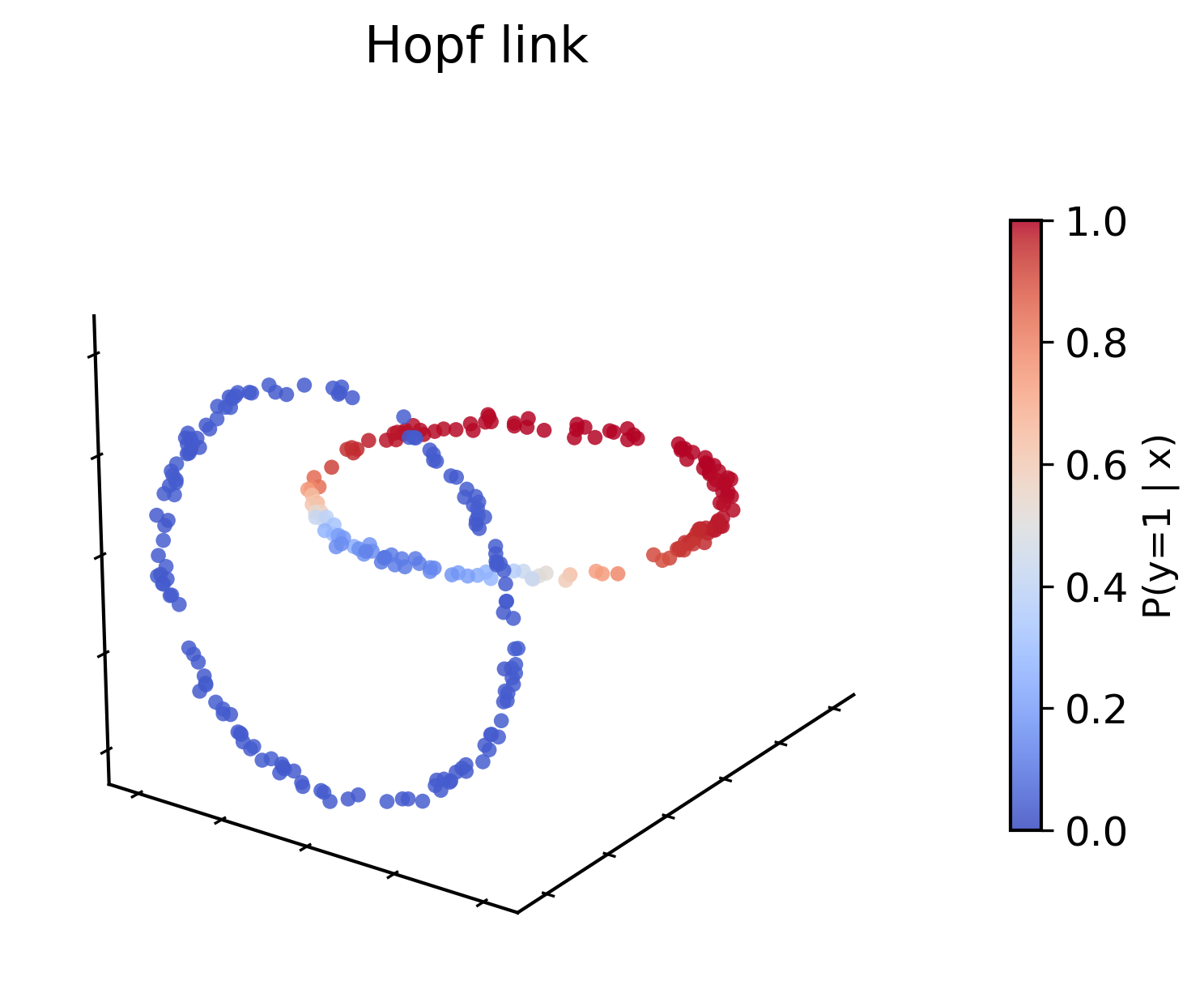}\caption{Hopf link}\end{subfigure}\hfill
\begin{subfigure}{0.31\textwidth}\includegraphics[width=\linewidth]{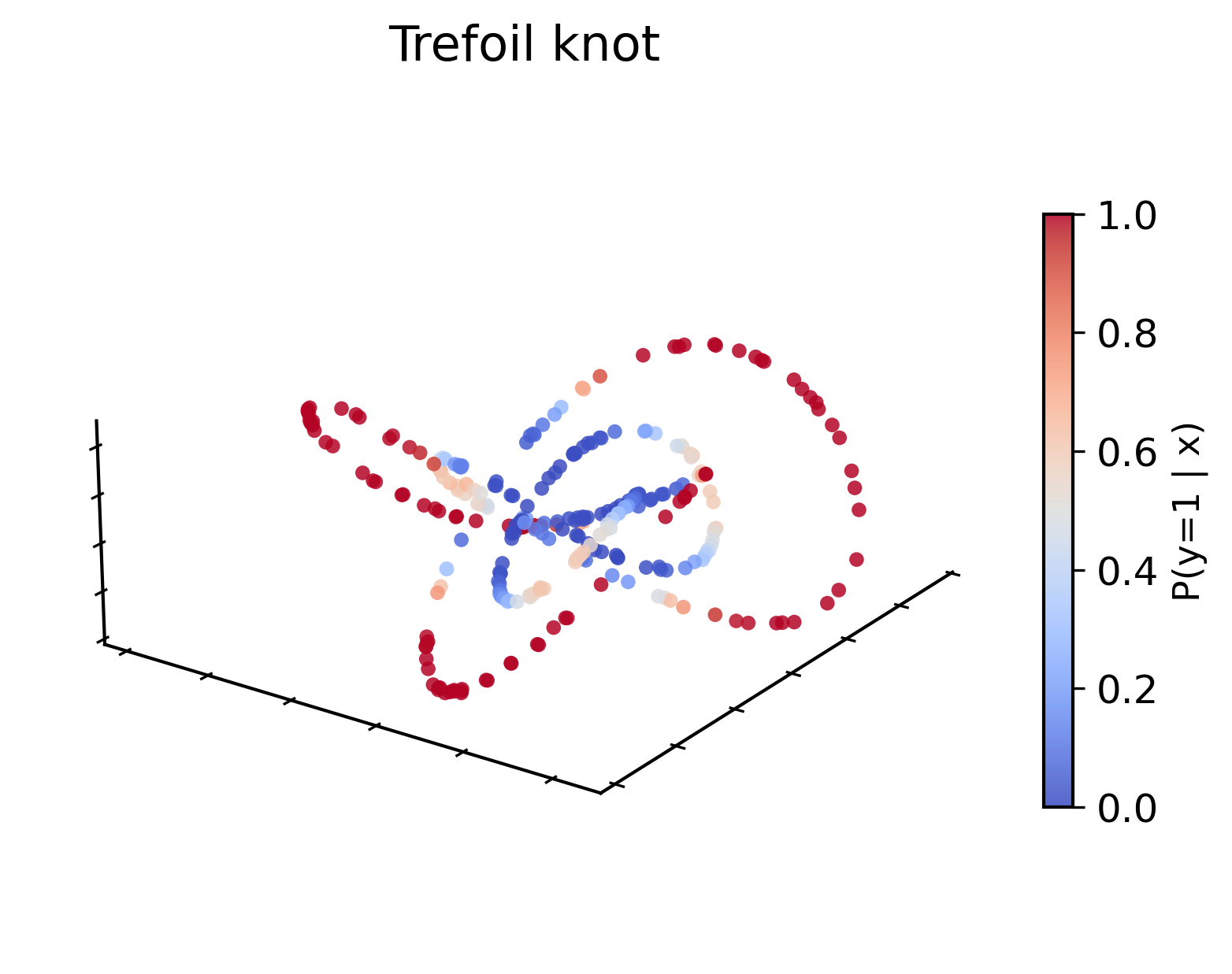}\caption{Trefoil knot}\end{subfigure}\hfill
\begin{subfigure}{0.31\textwidth}\includegraphics[width=\linewidth]{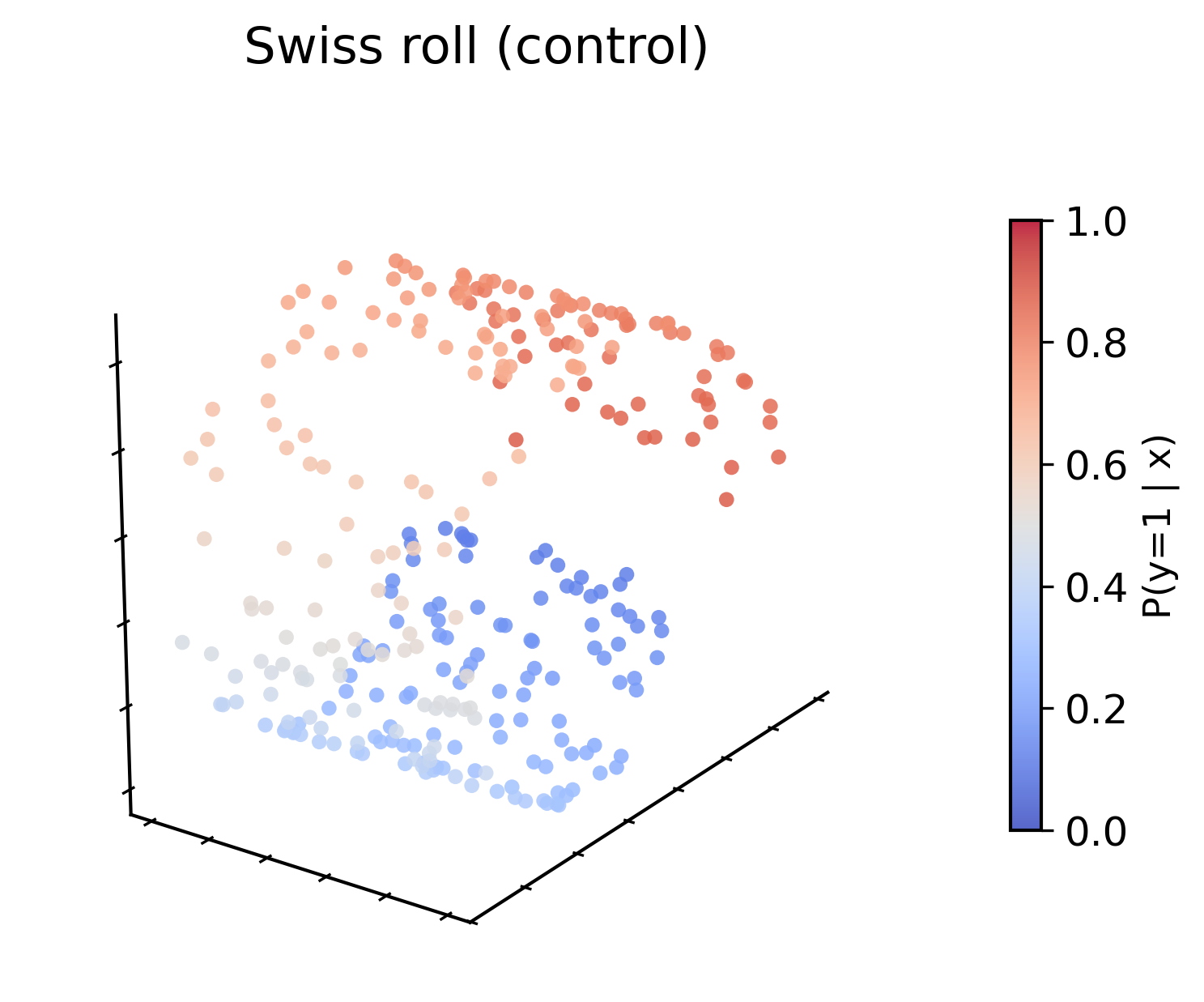}\caption{Swiss roll}\end{subfigure}
\caption{The six controlled topology families, shown as the ambient point cloud of a representative level 0 (lowest difficulty level) test set colored by the true probability $P(y{=}1\mid x)$. The warped circle is planar; the remaining families are embedded in $\mathbb{R}^3$.}
\label{fig:families}
\end{figure}

Table~\ref{tab:families} summarizes the six families and Figure~\ref{fig:families} visualizes them. The warped circle is a one-loop $S^1$ baseline; the torus is a two-loop $T^2$ structure; the sphere is a surface topology included to test whether $H_1$-focused descriptors are sufficient; the Hopf link consists of two interlinked circles; the trefoil knot is a knotted one-dimensional manifold; and the Swiss roll is a geometrically curved but topologically loop-free surface. The Swiss roll serves as a negative $H_1$ control where $H_1$ growth reflects spurious or representation-induced cycles rather than true input topology.

Each family is generated at multiple difficulty levels by increasing noise, nuisance dimensions, warping, curvature, or reducing separation. Label sharpness is held fixed to avoid conflating geometric difficulty with changing Bayes noise.

\begin{table}[t]
\centering
\caption{The six controlled topology families. Label sharpness is held fixed so that geometric difficulty is decoupled from Bayes noise.}
\label{tab:families}
\begin{tabular}{llcl}
\toprule
Family & Intrinsic topology & Expected $H_1$ & Role \\
\midrule
Warped circle & $S^1$ & 1 loop & Baseline \\
Torus & $T^2$ & 2 loops & Intermediate \\
Sphere & $S^2$ & 0 (surface) & $H_1$-insufficiency test \\
Hopf link & two linked $S^1$ & 2 linked loops & Linked loops \\
Trefoil knot & knotted $S^1$ & persistent $H_1$ & Entangled knot \\
Swiss roll & $\mathbb{R}^2$ sheet & $\approx 0$ & Negative $H_1$ control \\
\bottomrule
\end{tabular}
\end{table}

\subsection{Experimental scale}
The main topology-difficulty suite contains 180 runs:
\begin{equation}
6\ \text{families}\times 6\ \text{difficulty levels}\times 5\ \text{seeds},
\end{equation}
each generated with $n_{\mathrm{train}}=400$ context rows and $n_{\mathrm{test}}=300$ query rows. We also run an extreme suite at levels $6$--$8$ to probe saturation behavior under stronger noise and higher nuisance dimension.

To establish the core relationship at high resolution, we run a large-scale study on the warped circle, the family with the clearest single-loop topology and natural class balance, across all nine difficulty levels with $n_{\mathrm{train}}=6667$ and $n_{\mathrm{test}}=5000$. As in the main suite, we use five seeds per level, giving $45$ runs. We have also repeated the same design at half scale ($n_{\mathrm{train}}=3333$, $n_{\mathrm{test}}=2500$) to test scale-invariance which the results are briefly mentioned as patterns are similar to the large-scale run. We further run fine-tuning experiments and embedding-scope robustness checks to verify that the main observations are not artifacts of the default setup.

\section{Experimental Results}

We first establish the topology--reliability relationship at high resolution using the large-sample warped circle case study. We then return to the full six-family benchmark and show how the same pattern generalizes, where it breaks, and which parts of the signature remain stable across scale.

\subsection{Large-sample case study: the warped circle at scale}
\label{sec:hero}
To probe the core claim with minimal confounds, we study the warped circle ($S^1$) in depth at $n_{\mathrm{train}}=6667$ and $n_{\mathrm{test}}=5000$.
This family has a single interpretable $H_1$ feature and near-balanced labels by construction, so a difficulty ladder can change its geometry without changing the underlying homotopy type.

\begin{figure}[t]
\centering
\begin{subfigure}{0.32\textwidth}\includegraphics[width=\linewidth]{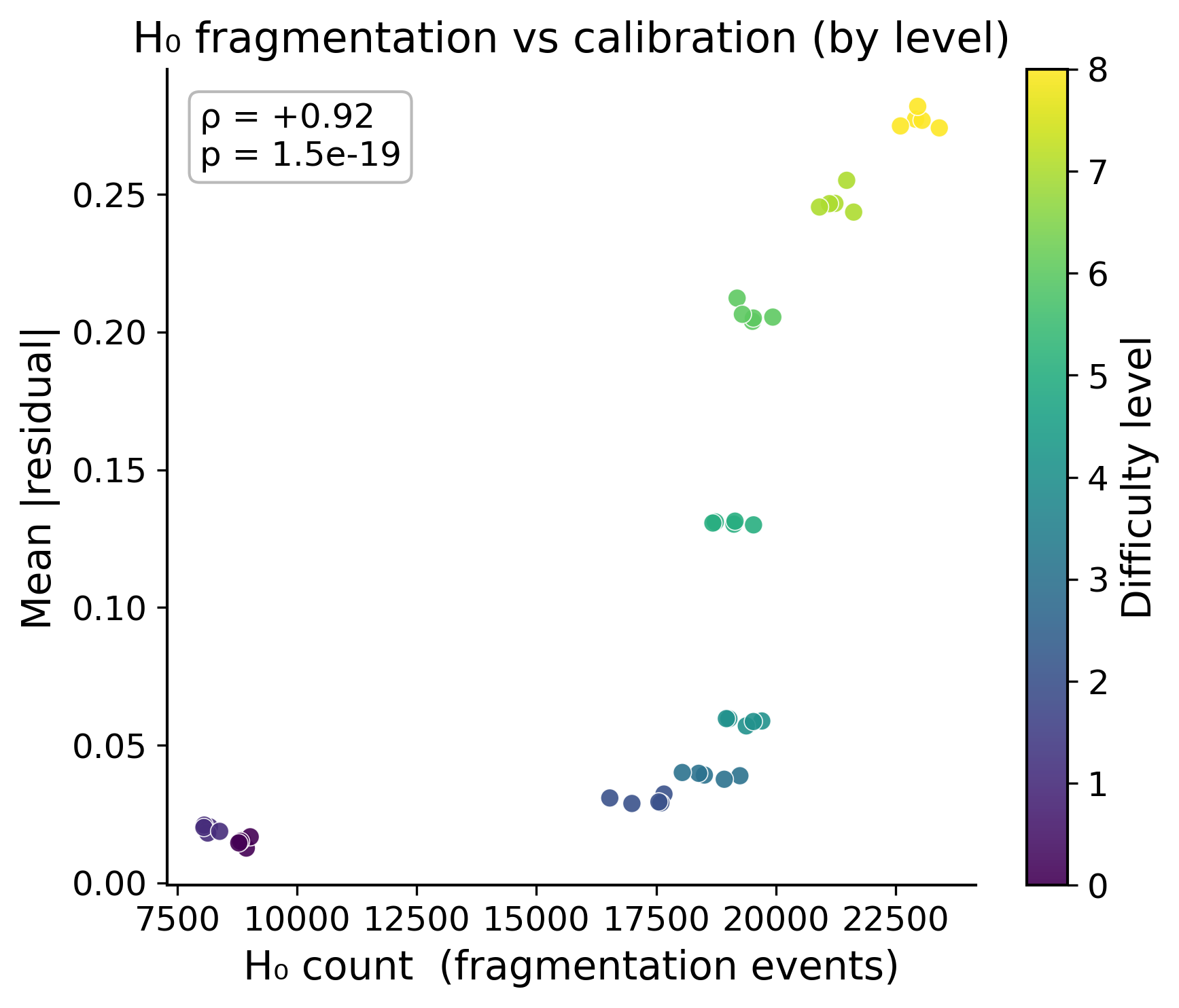}\caption{}\end{subfigure}\hfill
\begin{subfigure}{0.32\textwidth}\includegraphics[width=\linewidth]{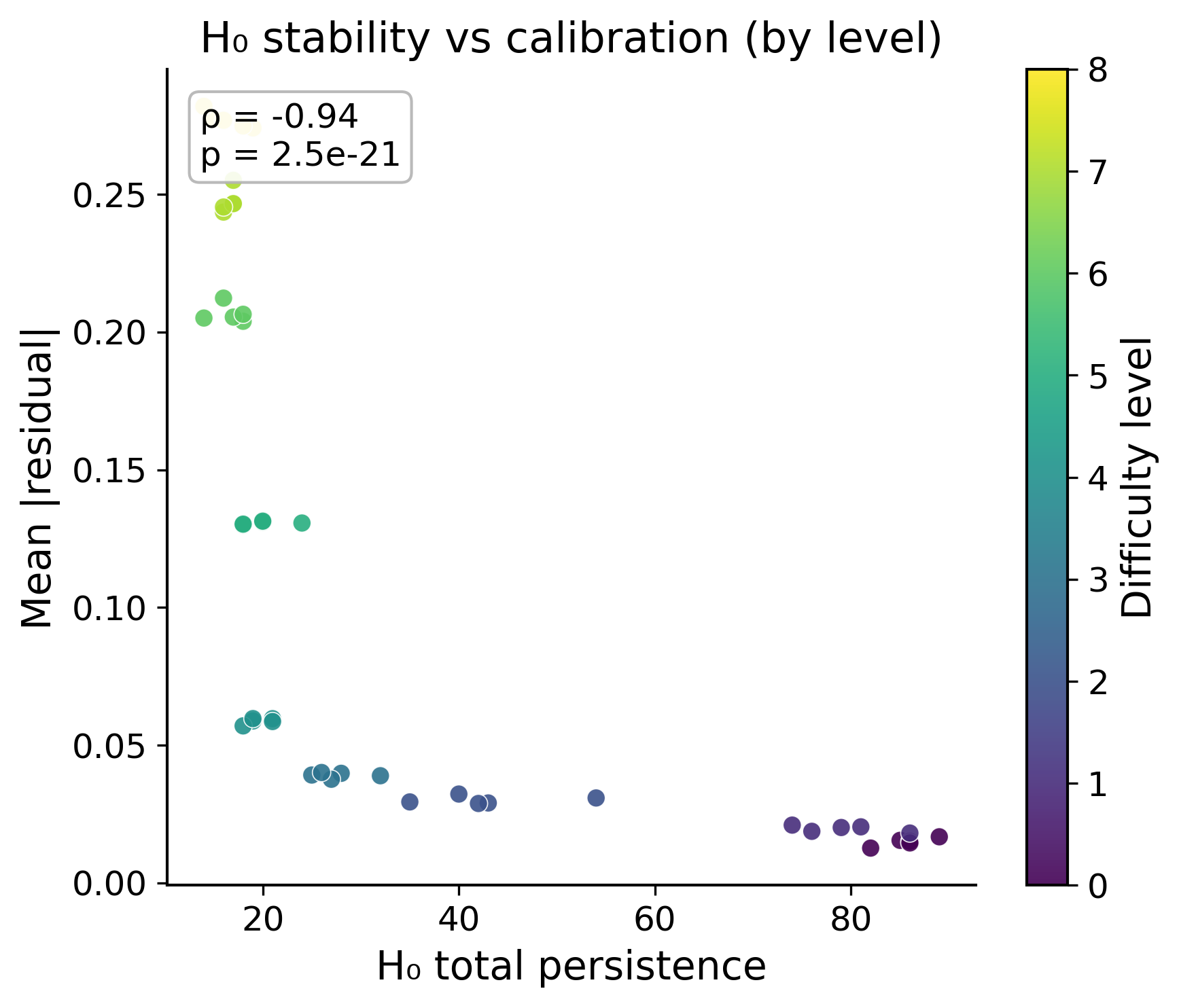}\caption{}\end{subfigure}\hfill
\begin{subfigure}{0.32\textwidth}\includegraphics[width=\linewidth]{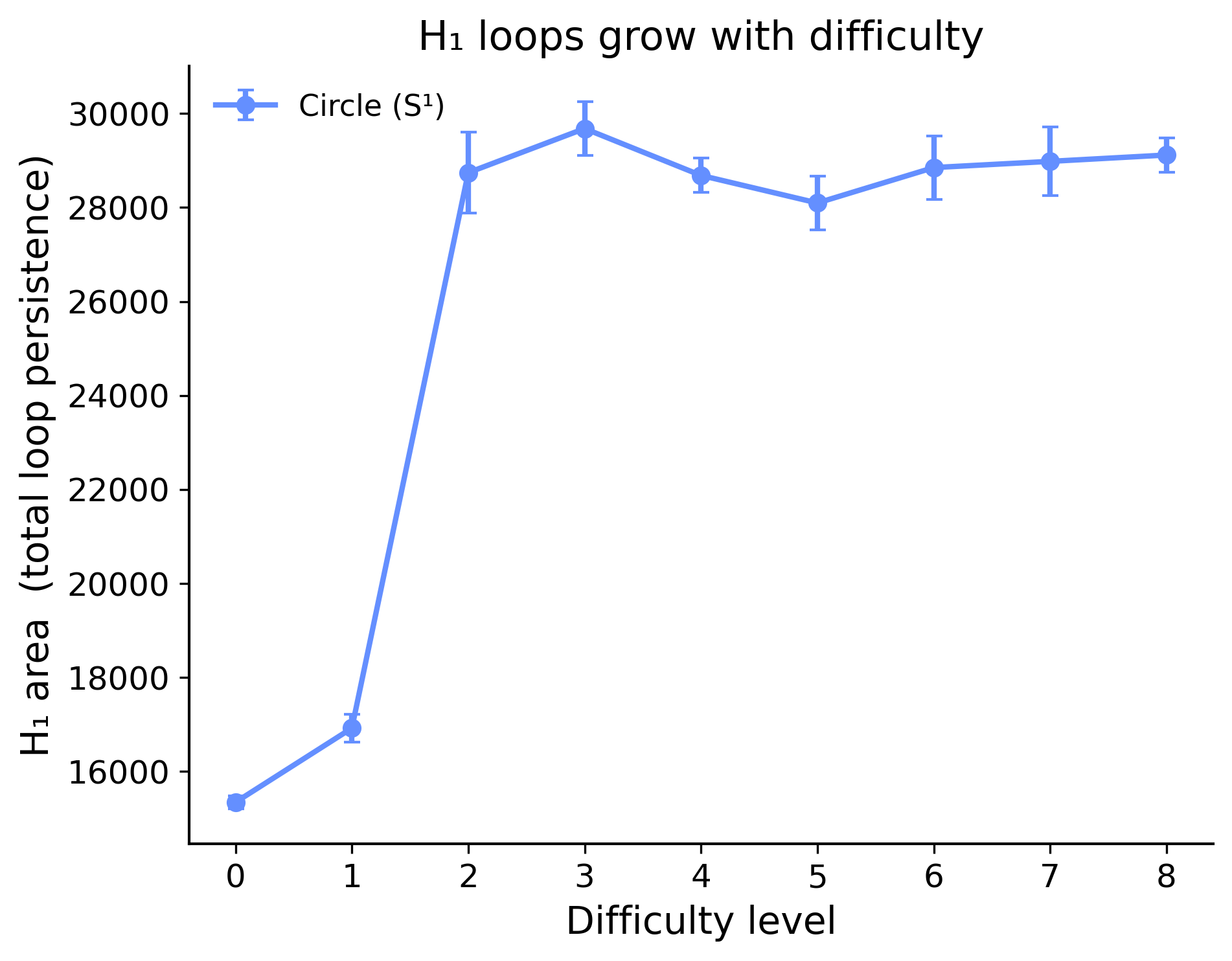}\caption{}\end{subfigure}
\caption{warped circle case study at scale ($45$ runs, $n_{\mathrm{test}}=5000$), with points in (a,b) colored by difficulty level. (a) $H_0$ fragmentation count vs MAR ($\rho=0.92$) and (b) $H_0$ total persistence vs MAR ($\rho=-0.94$): the color gradient shows that fragmentation moves monotonically with difficulty while simultaneously tracking calibration, so a single scatter conveys both the level progression and the reliability relationship. (c) $H_1$ area rises then saturates by level 2 once the loop is densely resolved, which is why raw $H_1$ area is a weaker large-sample predictor than the ceiling-free $H_0$ descriptors.}
\label{fig:hero}
\end{figure}

\paragraph{$H_0$ fragmentation strongly tracks reliability at scale.}
At scale, the $H_0$ fragmentation descriptors track true-probability reliability with strong monotone associations, as shown in Table~\ref{tab:hero} and Figure~\ref{fig:hero}. Pooled over all $45$ runs, the $H_0$ fragmentation count correlates with MAR at $\rho=0.92$ and with Bayes error at $\rho=0.94$, while $H_0$ total persistence correlates at $\rho=-0.94$ with both metrics. The relationship is not tied to a single error definition: the same fragmentation count also tracks observed error, overconfidence, and wrong-overconfidence. The durable $H_1$ signal points in the same direction from the loop side, with long-lived loop persistence falling as difficulty increases. At high resolution, the two homological dimensions agree on the same failure pattern: reliability degrades as the representation fragments.

\paragraph{$H_1$ area saturates at high sampling density.}
In contrast, raw $H_1$ area is a weaker predictor at scale. Figure~\ref{fig:hero}(c) shows that with $n_{\mathrm{test}}=5000$ densely sampled points, the circle's single loop is already resolved by difficulty level $2$, after which $H_1$ area plateaus even as error keeps climbing. The $H_0$ descriptors do not show the same ceiling. The fragmentation count rises monotonically from about $9000$ to more than $23000$, and total persistence falls across the full range. At high resolution, because the magnitude of $H_1$ saturates, the stress signal therefore shifts to the $H_0$ fragmentation and the loss of durable $H_1$ structure.

This saturation is expected. For a fixed neighborhood size $k$, the k-nearest-neighbor graph on $N$ points has a finite cycle capacity, on the order of $(k-1)N$ independent $1$-cycles. At high density, even mild difficulty can push the graph close to that capacity, compressing the dynamic range of raw loop counts and areas. Consistent with this, $H_1$ count rises by 3 times across levels at $n_{\mathrm{test}}=300$, but only by less than 2 times at $n_{\mathrm{test}}=5000$. Persistent homology also stabilizes as dense samples converge to the underlying manifold, so the large finite-sample swings in $H_1$ area at small $n$ should not be expected to persist indefinitely \cite{cohen2007stability,niyogi2008finding}. The $H_0$ fragmentation count and durable-$H_1$ persistence have no comparable raw-area ceiling, and they continue to change monotonically with difficulty. This hints why $H_0$ fragmentation is the more scale-stable reliability signal.

\paragraph{The hard-versus-easy contrast is extreme.}
Splitting the case study into its three easiest and three hardest levels, Bayes error rises from 0.015 to 0.223, and MAR rises from 0.022 to 0.244. The same split shows $H_0$ total persistence collapsing from 69 to 17 and the $H_0$ fragmentation count nearly doubling. These are among the largest standardized effects we observe, helped by the low variance that dense sampling affords.

This case study gives the least confounded version of the paper's central relationship. Once sampling noise is reduced, $H_0$ fragmentation and durable-$H_1$ collapse become strong readouts of where TabPFN's predictions become unreliable. The descriptor that weakens, raw $H_1$ area, does so for a concrete reason: the loop has already been resolved, so the raw area has little room left to grow.

\begin{table}[t]
\centering
\caption{warped circle case study at scale ($n=45$, $n_{\mathrm{test}}=5000$). Pooled Spearman correlations between embedding topology and true-probability reliability. $H_0$ fragmentation is strongly associated with reliability; raw $H_1$ area saturates once the loop is densely resolved.}
\label{tab:hero}
\begin{tabular}{lcc}
\toprule
Descriptor & vs.\ MAR & vs.\ Bayes error \\
\midrule
$H_0$ total persistence & $-0.94$ & $-0.94$ \\
$H_0$ fragmentation count & $0.92$ & $0.94$ \\
Durable $H_1$ persistence & $-0.86$ & $-0.87$ \\
$H_1$ area (saturates) & $0.52$ & $0.54$ \\
\bottomrule
\end{tabular}
\end{table}

\subsection{The pattern generalizes across six topology families}
\label{sec:systematic}
\begin{figure}[t]
\centering
\begin{subfigure}{0.49\textwidth}\includegraphics[width=\linewidth]{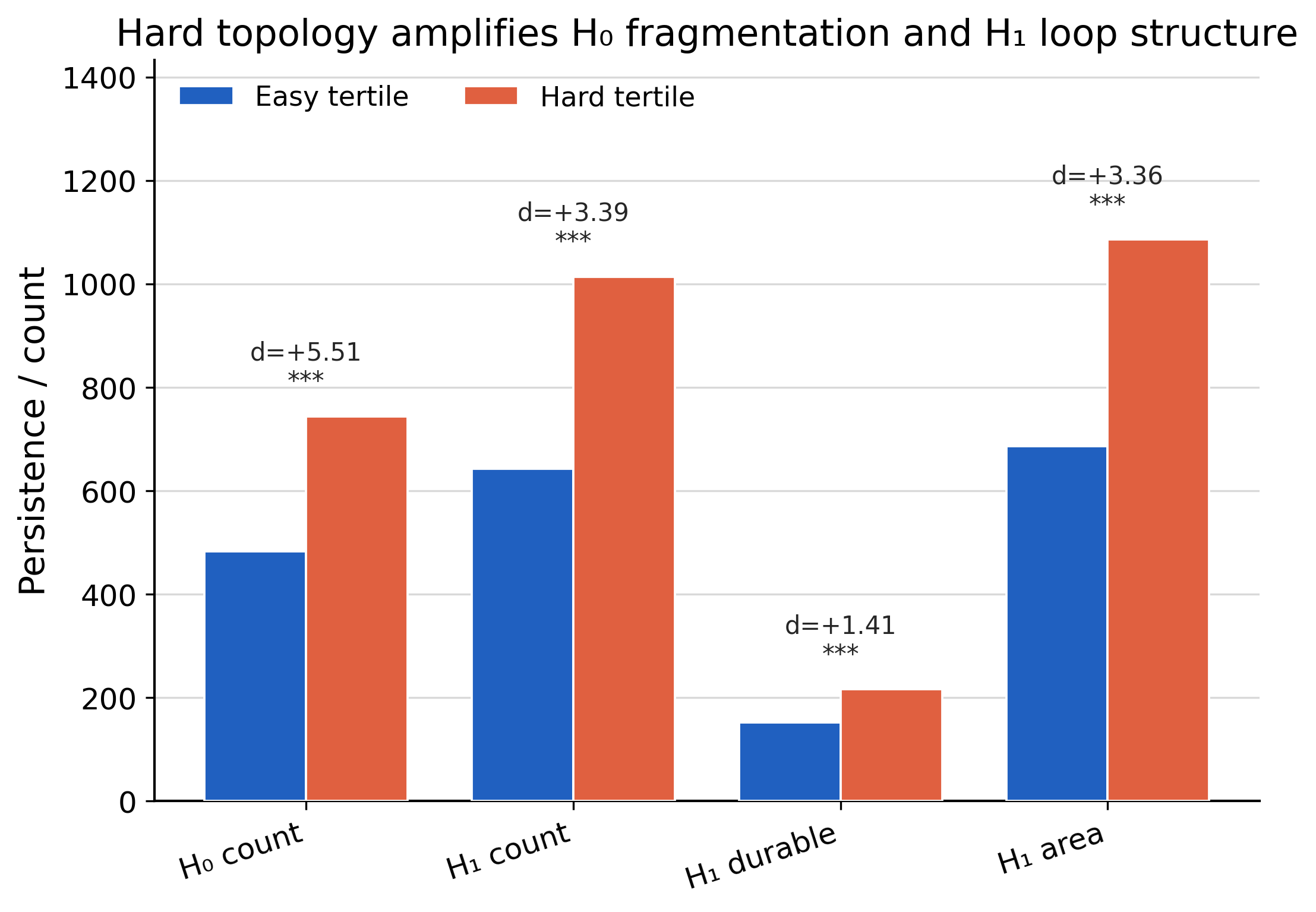}\caption{}\end{subfigure}\hfill
\begin{subfigure}{0.49\textwidth}\includegraphics[width=\linewidth]{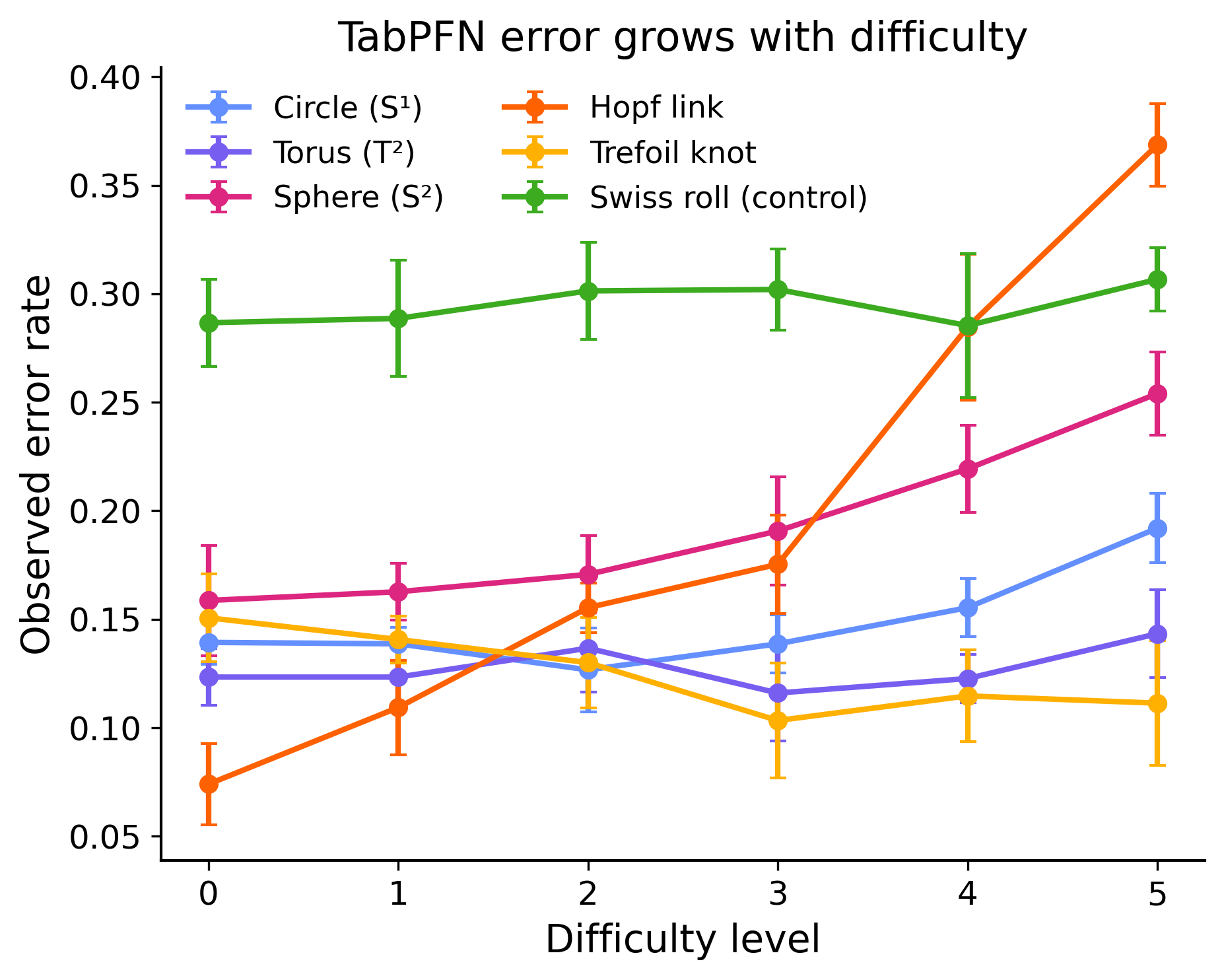}\caption{}\end{subfigure}
\caption{(a) Embedding-topology descriptors in the easy vs hard difficulty tertiles; all loop and fragmentation-count metrics increase under topological stress (Cohen's $d$ and Mann--Whitney significance annotated). (b) TabPFN's observed error grows with difficulty level in the main topology-difficulty suite across all six families.}
\label{fig:systematic}
\end{figure}

Across the full six-family benchmark, Figure~\ref{fig:systematic} indicates that TabPFN's internal topology changes systematically as the input geometry becomes more difficult. Comparing the easiest and hardest difficulty tertiles, split by ground-truth difficulty level, the loop- and fragmentation-related descriptors all increase with topological stress: $H_1$ area grows from 686 to 1086, the number of $H_1$ loops from 643 to 1014, durable $H_1$ persistence from 152 to 216, and the $H_0$ fragmentation count from 484 to 744. TabPFN's observed error rises across the same difficulty range as seen in Figure~\ref{fig:systematic}b. The $H_0$ fragmentation count has the largest hard-versus-easy effect among the reported topology descriptors, exceeding the loop-activity metrics and matching the case-study result that fragmentation is the dominant signal. Model reliability shifts in the same direction: MAR, Bayes error, and the wrong-overconfidence rate all rise.

A complementary movement appears in $H_0$ total persistence, which decreases as difficulty grows. The representation graph produces more connected-component events, but each is shorter-lived: the mean $H_0$ bar lifetime falls. This is the first sign of the $H_0/H_1$ scissors developed in Section~\ref{sec:scissors}: the task representation becomes more eventful, but less stable.

Together, these results show that the large-sample warped circle relationship is not a peculiarity of the circle. Across linked, knotted, surface, and control geometries, harder inputs push TabPFN toward a more fragmented and more cyclic internal representation, and that movement coincides with a measurable loss of reliability.

\subsection{$H_0$ and $H_1$ predict different reliability axes}
\label{sec:axes}
\begin{figure}[t]
\centering
\begin{subfigure}{0.49\textwidth}\includegraphics[width=\linewidth]{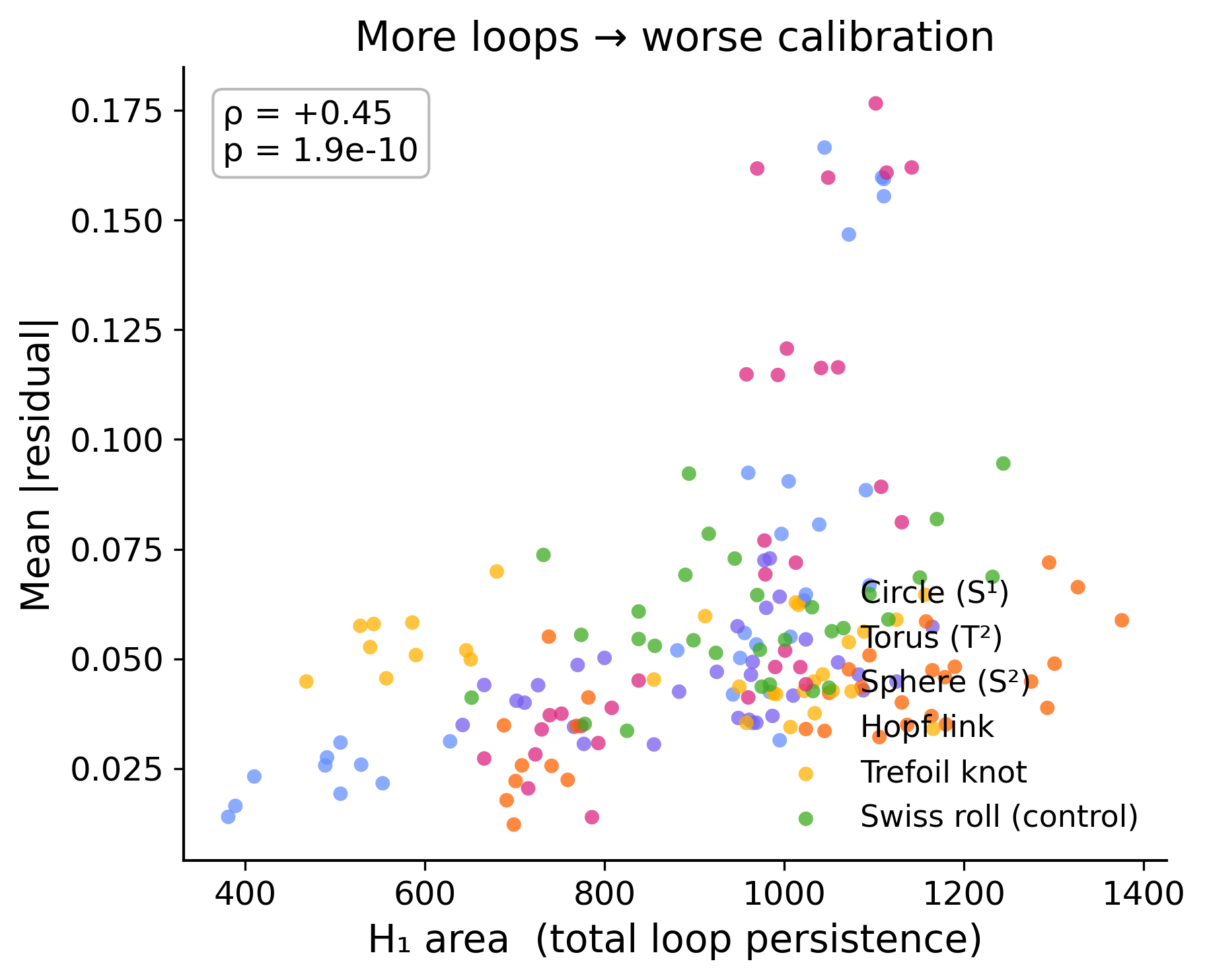}\caption{}\end{subfigure}\hfill
\begin{subfigure}{0.49\textwidth}\includegraphics[width=\linewidth]{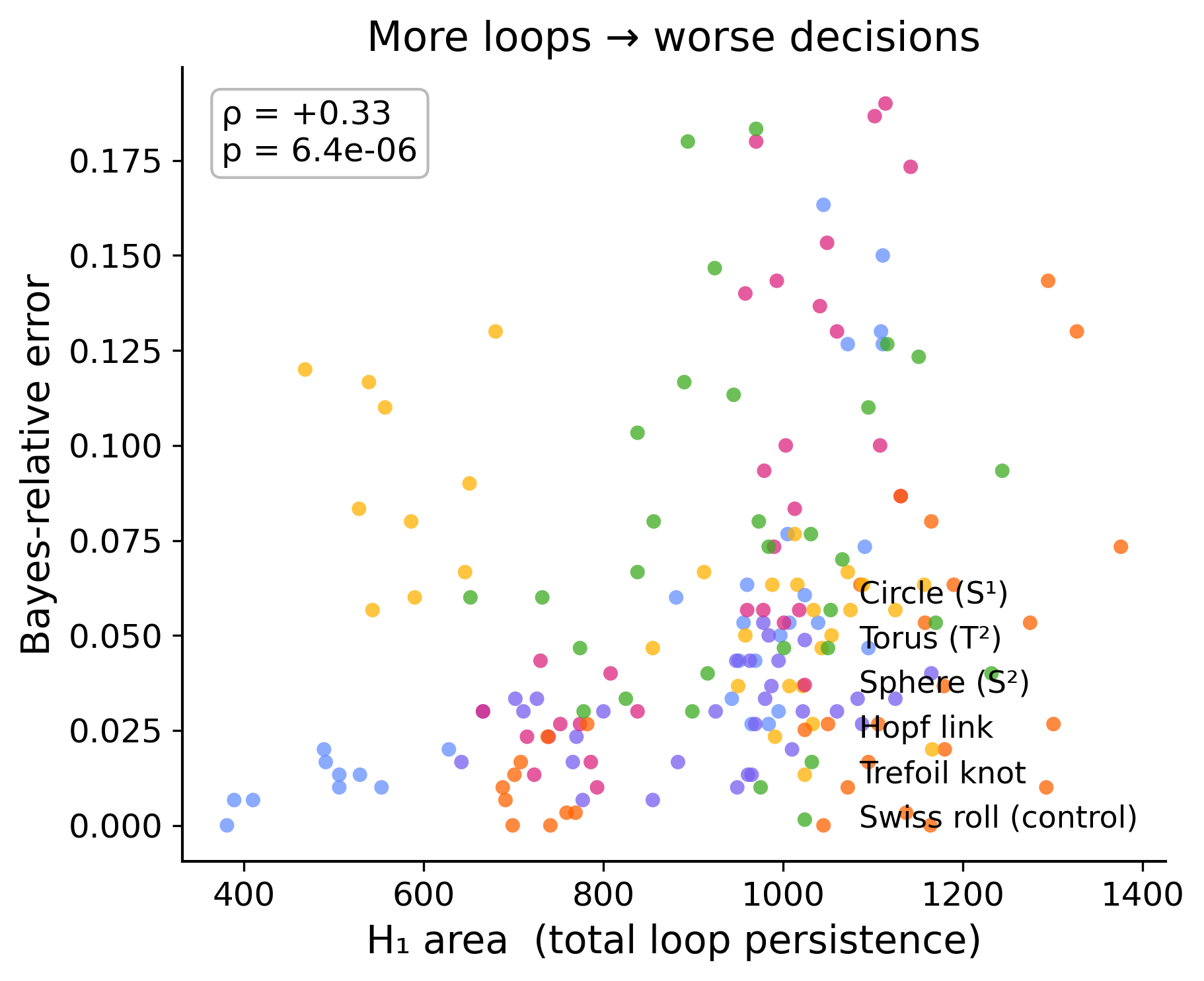}\caption{}\end{subfigure}

\medskip
\begin{subfigure}{0.49\textwidth}\includegraphics[width=\linewidth]{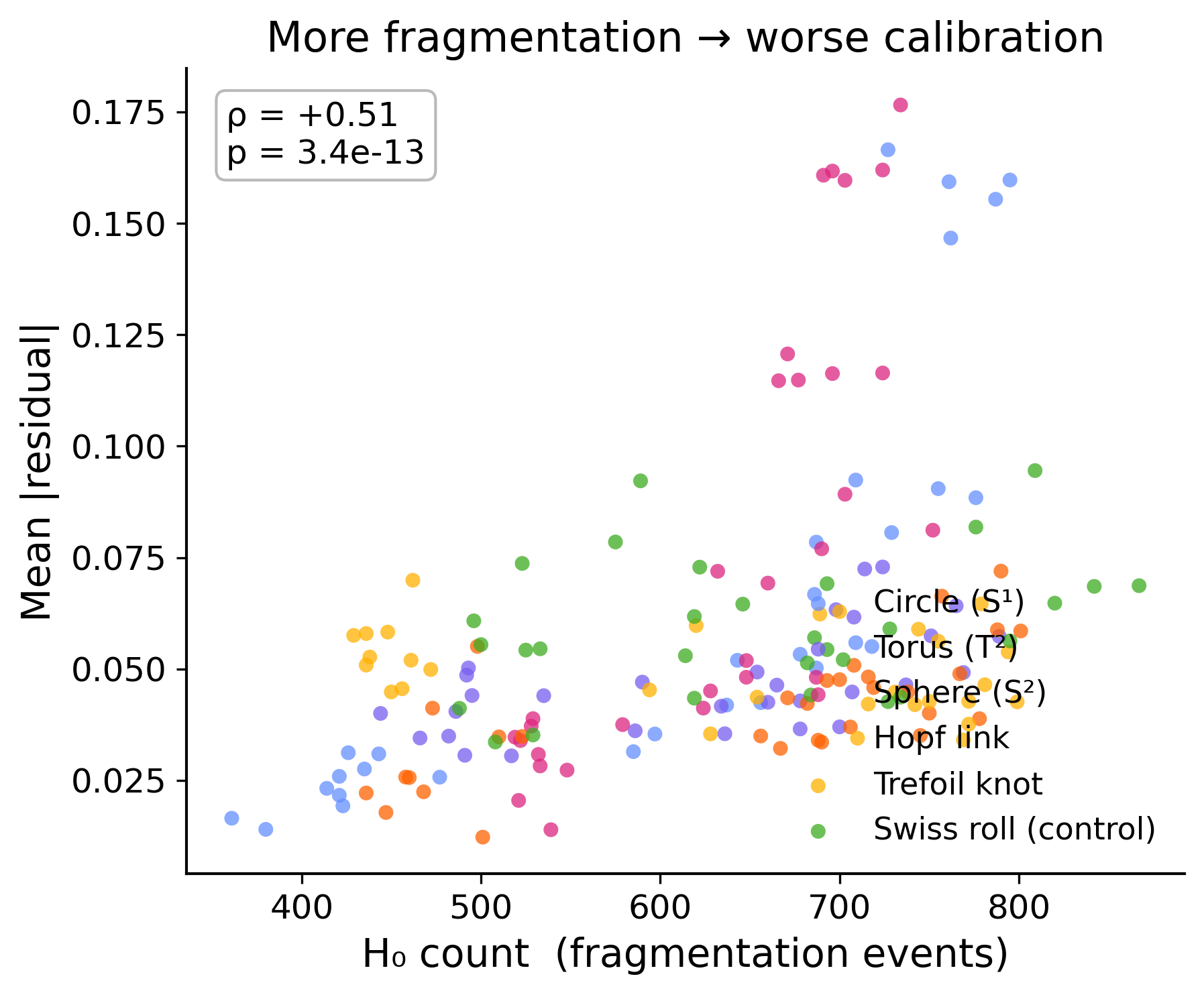}\caption{}\end{subfigure}\hfill
\begin{subfigure}{0.49\textwidth}\includegraphics[width=\linewidth]{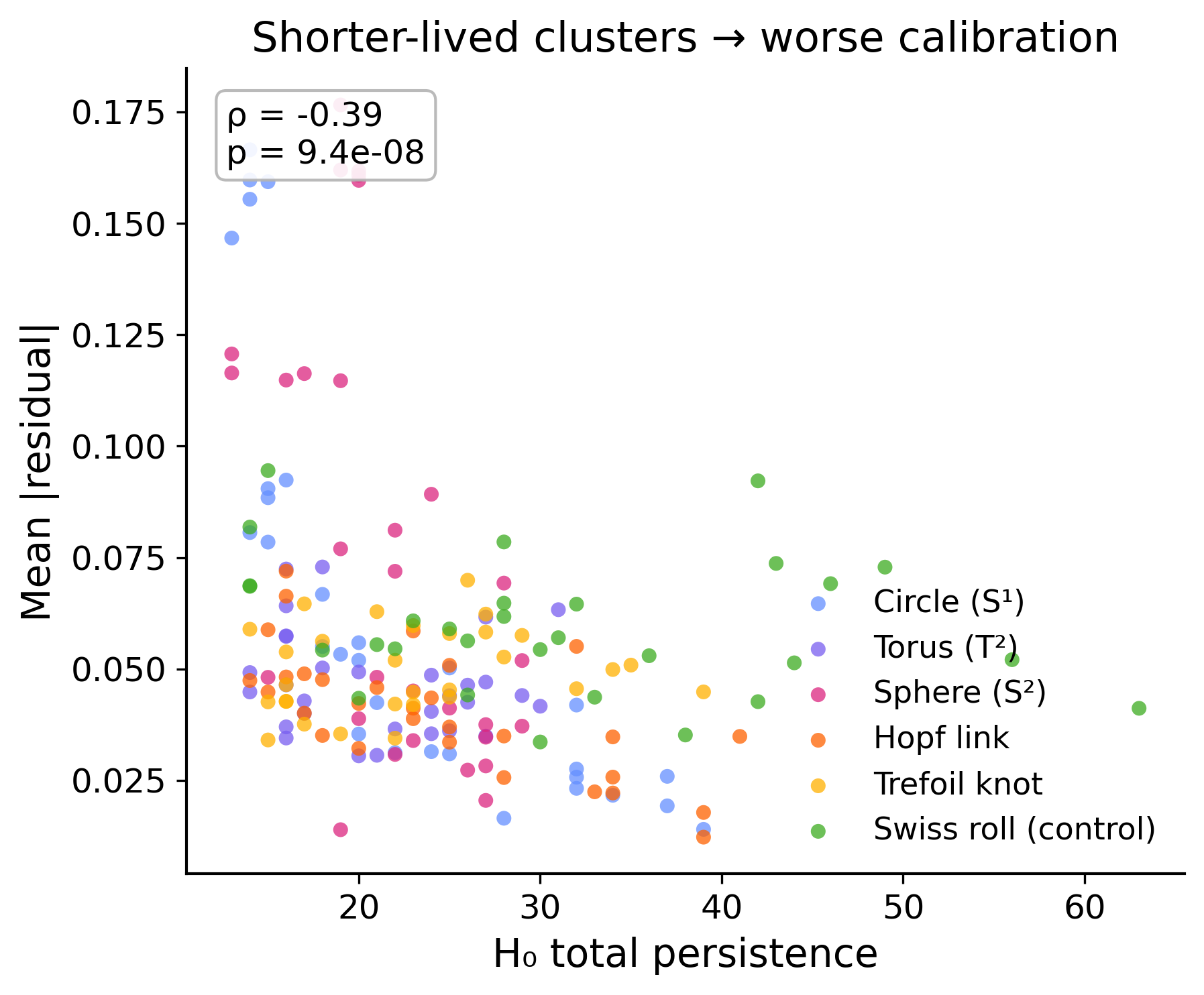}\caption{}\end{subfigure}
\caption{Dataset-level relationships between embedding topology and reliability (each point is one of the $180$ runs, colored by family). (a) $H_1$ area vs MAR; (b) $H_1$ area vs Bayes error; (c) $H_0$ fragmentation count vs MAR, the strongest single predictor; (d) $H_0$ total persistence vs MAR, showing that shorter-lived clusters accompany worse calibration. Inset annotations report the pooled Spearman $\rho$.}
\label{fig:scatters}
\end{figure}

The topological descriptors correlate strongly with true-probability reliability metrics presented in Table~\ref{tab:corr} and Figure~\ref{fig:scatters}. Pooled across all $180$ runs, $H_1$ area correlates positively with MAR, observed error, and Bayes error. Loop activity is therefore not only a geometric artifact: it tracks datasets on which both decisions and probabilities worsen.

\begin{figure}[t]
\centering
\includegraphics[width=0.72\textwidth]{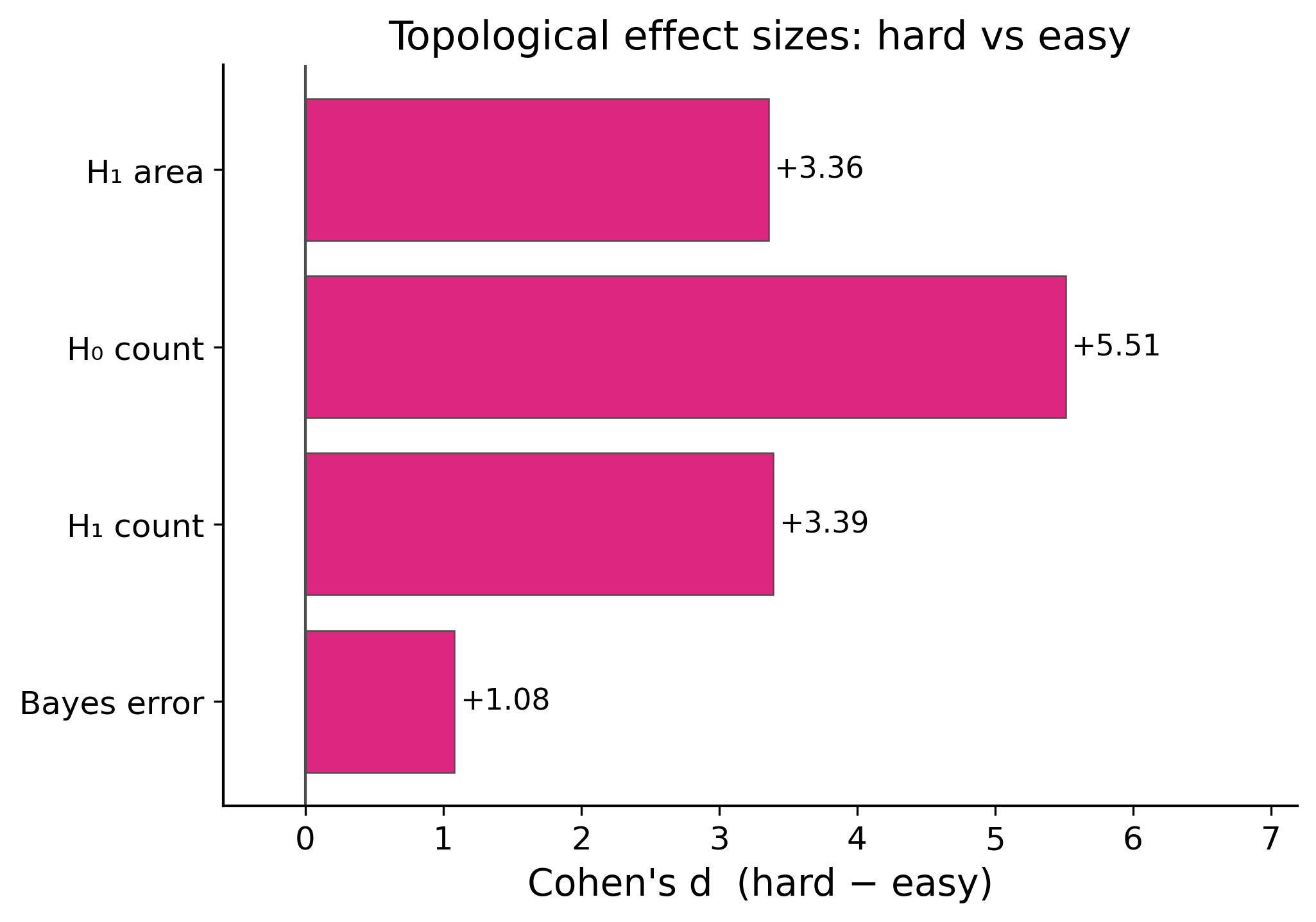}
\caption{Standardized effect sizes (Cohen's $d$) for the hard-versus-easy tertile contrast. Loop activity and the $H_0$ fragmentation count rise with difficulty while Bayes error increases; $H_0$ total persistence (Section~\ref{sec:scissors}) moves in the opposite direction.}
\label{fig:effectsizes}
\end{figure}

The $H_0$ descriptors are especially informative for calibration. The $H_0$ fragmentation count is the single strongest predictor of MAR, whereas $H_0$ total persistence and $H_0$ early-layer mass decrease as reliability worsens. Calibration failure is therefore tied to fragmentation and to the loss of stable, early-formed grouping structure, echoing the $H_0$ dominance seen in the high-resolution case study in Section~\ref{sec:hero}. Figure~\ref{fig:effectsizes} ranks the standardized effect sizes of the hard-versus-easy contrast.

The two dimensions also separate by failure type in the reported metrics. $H_1$ area is positively associated with MAR, Bayes error, and observed error, while early $H_0$ mass is the strongest predictor of Bayes error in Table~\ref{tab:corr}.
Thus $H_0$ and $H_1$ are not redundant; they capture complementary aspects of TabPFN's internal task geometry. $H_0$ tracks coarse organization and fragmentation, while $H_1$ tracks loop complexity.

\begin{table}[t]
\centering
\caption{Pooled Spearman correlations between embedding-topology descriptors and true-probability reliability metrics. Larger values of every reliability metric indicate worse reliability. ${}^{*}p<0.05$, ${}^{**}p<0.01$, ${}^{***}p<0.001$.}
\label{tab:corr}
\begin{tabular}{lccc}
\toprule
Descriptor & MAR & Bayes error & Observed error \\
\midrule
$H_1$ area & $0.45^{***}$ & $0.33^{***}$ & $0.34^{***}$ \\
$H_0$ count & $0.51^{***}$ & $0.37^{***}$ & $0.25^{***}$ \\
$H_0$ total & $-0.39^{***}$ & $-0.28^{***}$ & $-0.06$ \\
$H_0$ early frac. & $-0.41^{***}$ & $-0.42^{***}$ & $-0.35^{***}$ \\
\bottomrule
\end{tabular}
\end{table}

\subsection[The H0/H1 scissors effect]{The $H_0/H_1$ scissors effect}
\label{sec:scissors}
\begin{figure}[t]
\centering
\begin{subfigure}{0.32\textwidth}\includegraphics[width=\linewidth]{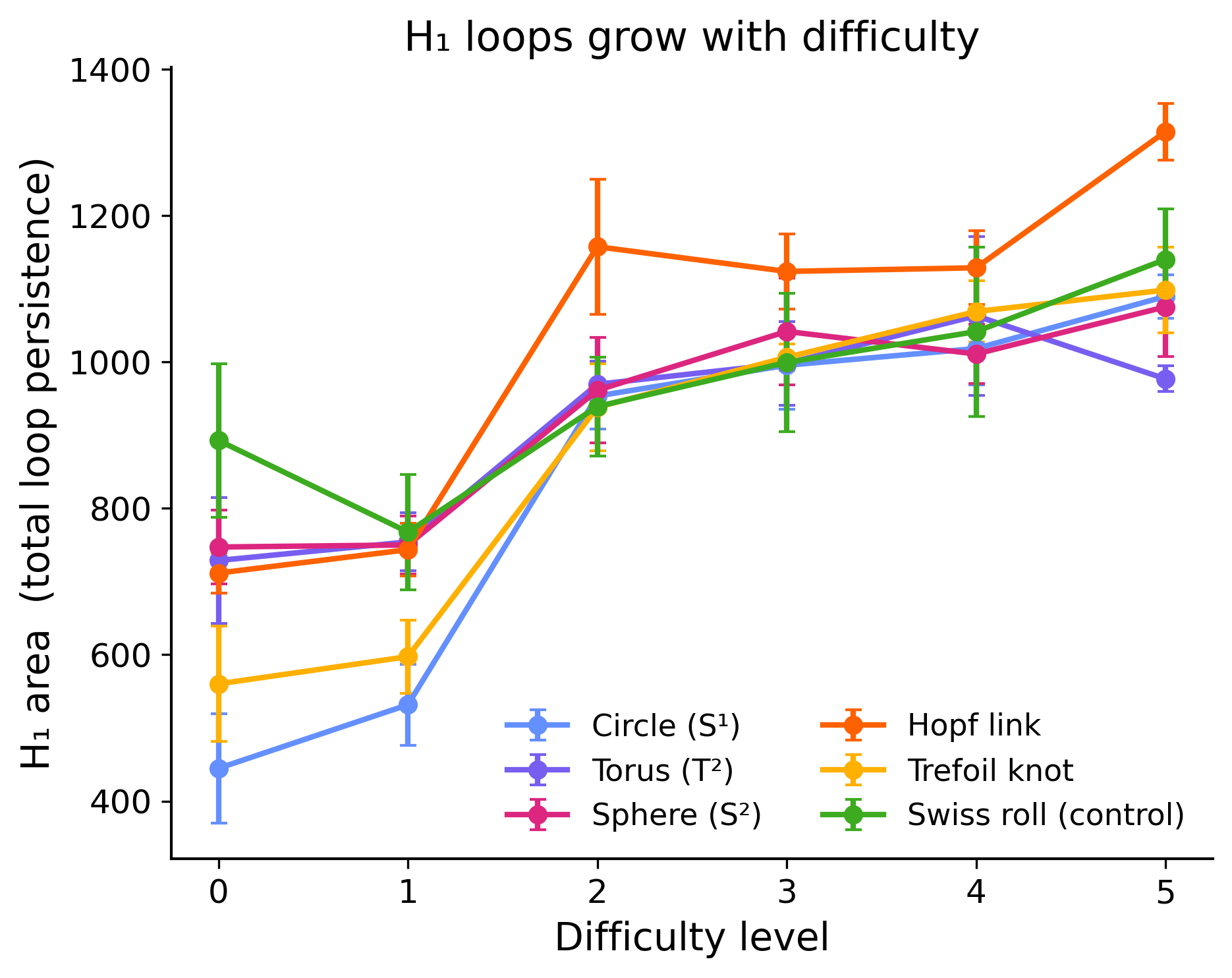}\caption{}\end{subfigure}\hfill
\begin{subfigure}{0.32\textwidth}\includegraphics[width=\linewidth]{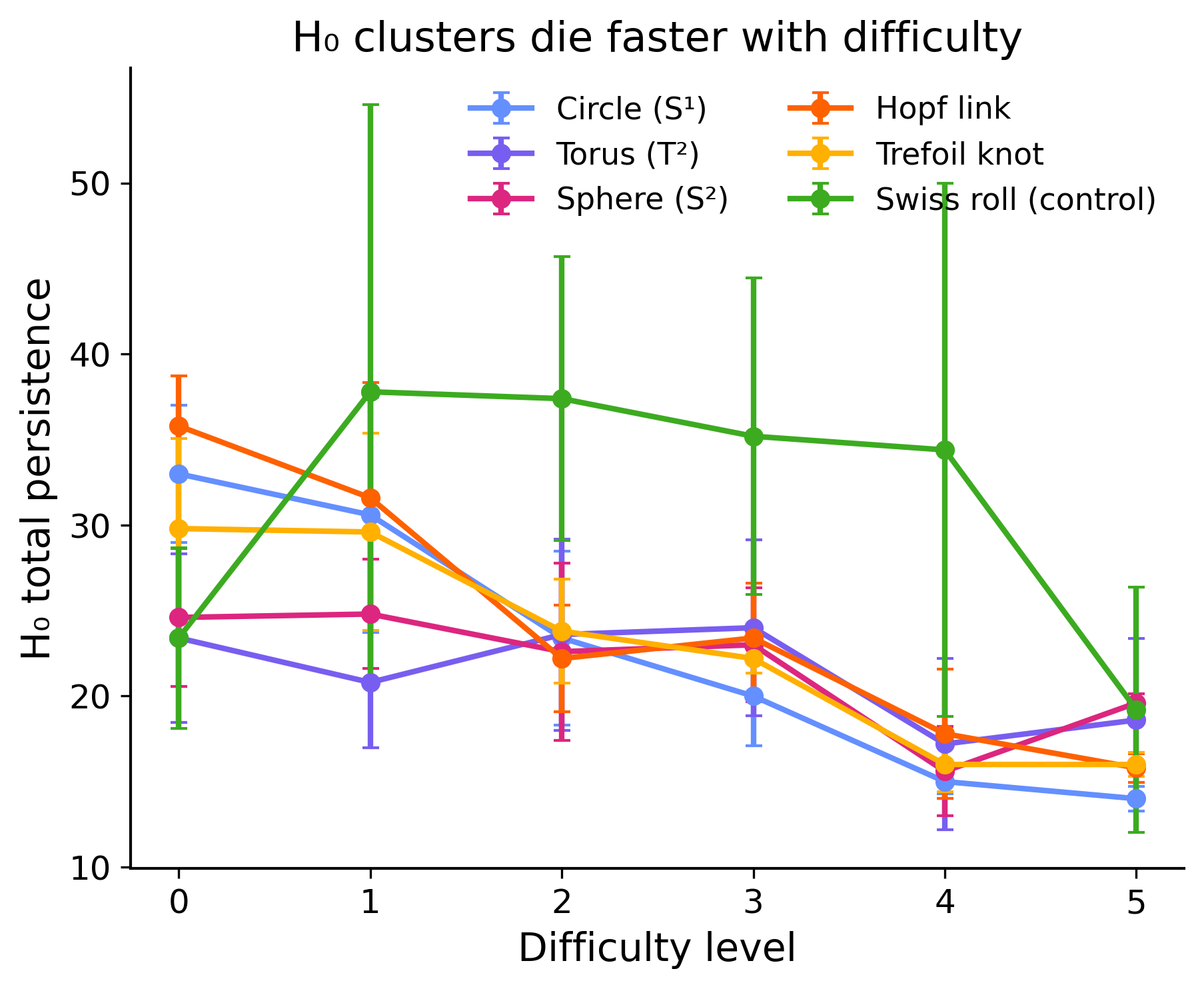}\caption{}\end{subfigure}\hfill
\begin{subfigure}{0.32\textwidth}\includegraphics[width=\linewidth]{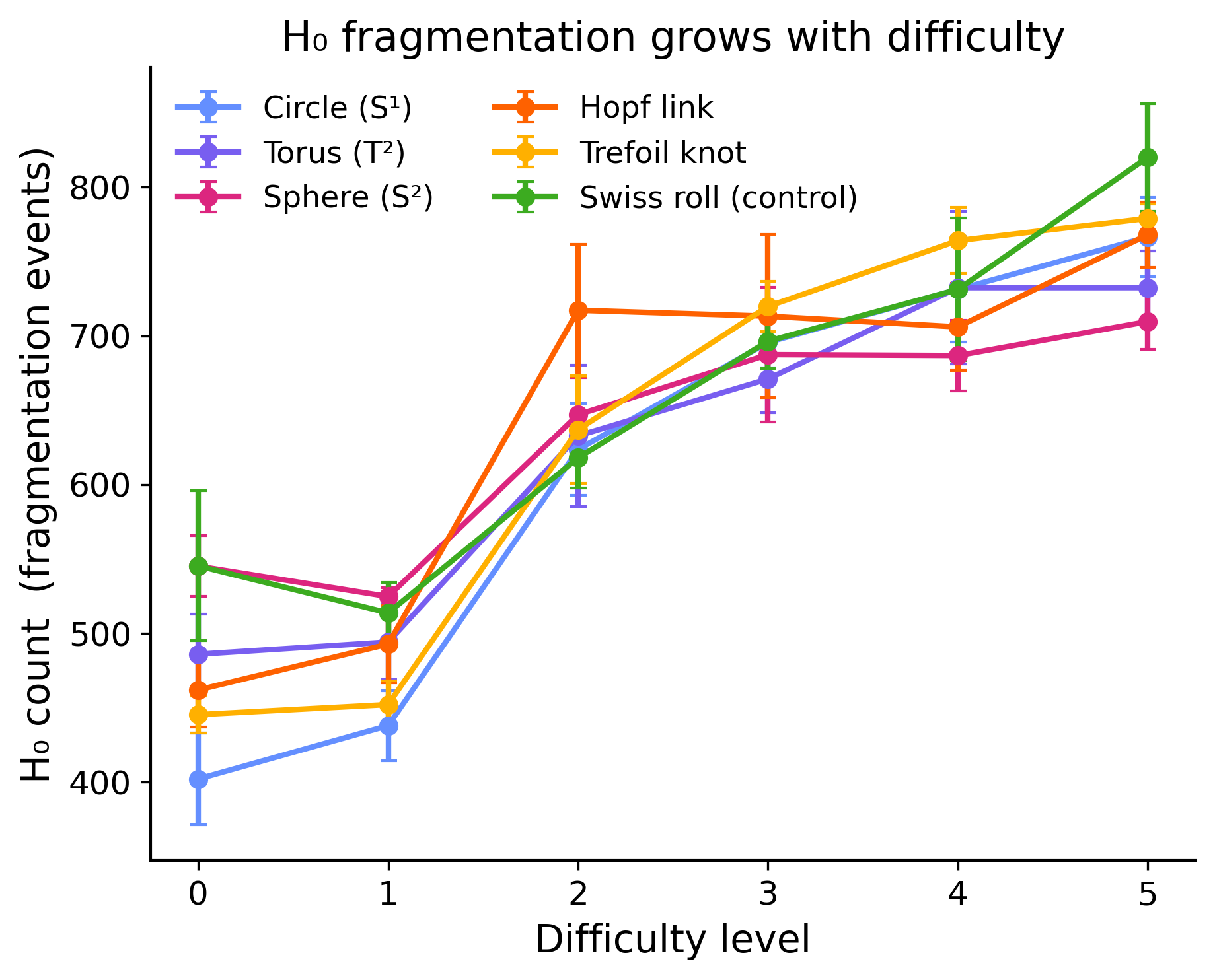}\caption{}\end{subfigure}
\caption{The $H_0/H_1$ scissors across difficulty levels (mean $\pm$ s.d.\ per family). (a) $H_1$ area rises with difficulty; (b) $H_0$ total persistence falls; (c) $H_0$ count rises. Loop structure proliferates while stable cluster structure fragments.}
\label{fig:scissors}
\end{figure}

The most interpretable pattern is the $H_0/H_1$ scissors effect seen in Figure~\ref{fig:scissors}. As difficulty increases, $H_1$ area grows while $H_0$ total persistence falls; for example, in the warped circle, $H_1$ area grows by roughly $2.4\times$ from level 0 to level 5, while $H_0$ total persistence falls from 33 to 14. The $H_0$ count rises at the same time. In words, loop structure proliferates while stable connected-component structure erodes.

\paragraph{A unified fragmentation principle.}
The high-resolution case study in Section~\ref{sec:hero} and a matched half-scale run ($n_{\mathrm{test}}=2500$) show that the scissors is, more fundamentally, a fragmentation principle acting in both homological dimensions: under stress the model produces more features, but they survive for less time. Table~\ref{tab:crossscale} reports how each descriptor correlates with difficulty level as sampling density grows from $n_{\mathrm{test}}=300$ to 5000. Feature counts rise with difficulty in both dimensions at every scale. Durable structure is more revealing: $H_0$ total persistence is negatively correlated with difficulty at all scales, and durable $H_1$ persistence reverses sign as resolution increases. At low sampling density, noisy short-lived bars inflate the apparent magnitude of loop activity; at high resolution, harder topology still produces more loop events, but those loops become less stable.

The durable-$H_1$ sign flip comes from two competing effects. Difficulty increases the number of cycles while shortening their lifetimes, so the fraction of loop persistence carried by durable bars falls at every scale, from 0.26 to 0.20 at $n_{\mathrm{test}}=300$ and from 0.35 to 0.11 at $n_{\mathrm{test}}=5000$. Whether the absolute durable mass rises or falls depends on which effect wins. At small $n$, the graph is far from its cycle ceiling and the count inflates nearly threefold across levels, pulling even the durable component upward. At large $n$, total growth is capped by saturation while the durable fraction collapses, so durable mass falls. Dense sampling builds a genuinely long-lived loop that noise can then destroy; sparse sampling never builds one as cleanly, so noise mostly inflates cycle counts. This is why the sign of the durable trend, but not the underlying fragmentation, depends on resolution.

TabPFN is not simply creating more topology in a uniform way. Under topological stress it fragments its representation in both dimensions: durable coarse organization in $H_0$ collapses, and durable relational structure in $H_1$ collapses as well. Raw $H_1$ area rises only until the manifold is densely resolved, after which it saturates. The scale-invariant signature of stress is therefore the proliferation of short-lived features together with the loss of durable structure.

\begin{table}[t]
\centering
\caption{The scissors is a scale-invariant fragmentation principle. Spearman $\rho$ between each descriptor and difficulty level for the warped circle (levels 0--8, $n=45$ runs per sample size). Feature counts rise with difficulty in both dimensions at every scale; durable $H_1$ persistence reverses sign as resolution increases, revealing that loops fragment just as connected components do.}
\label{tab:crossscale}
\begin{tabular}{lccc}
\toprule
$\rho(\cdot,\ \text{difficulty level})$ & $n_{\mathrm{test}}{=}300$ & $2500$ & $5000$ \\
\midrule
$H_0$ count & $0.98$ & $0.96$ & $0.94$ \\
$H_1$ count & $0.93$ & $0.77$ & $0.65$ \\
$H_1$ area & $0.95$ & $0.75$ & $0.53$ \\
$H_0$ total persistence (durable) & $-0.91$ & $-0.94$ & $-0.94$ \\
Durable $H_1$ persistence & $0.79$ & $-0.77$ & $-0.87$ \\
\bottomrule
\end{tabular}
\end{table}

\subsection{Trefoil knot reveals a second failure mode}
\label{sec:trefoil}
The per-family analysis in Table~\ref{tab:perfamily} shows that the topology--error relationship, while positive for five of six families, inverts for the trefoil knot. The trefoil generator uses a per-dataset median split, so class balance does not drift with difficulty. The inversion therefore reflects a different representational behavior: as the knot becomes harder to resolve, TabPFN appears to simplify or collapse its representation instead of generating more loop structure.

This pattern points to a second failure mode. For most topology families, reduced reliability is associated with increased topological complexity. For the trefoil, reduced reliability instead appears as representational collapse: as the knot becomes harder to resolve, the model simplifies the entangled structure rather than generating richer loop geometry. The collapse registers in both homological dimensions. The $H_0$ fragmentation count, which is positively coupled to error for the other structured families, also inverts for the trefoil. The inversion is strongest for decision-error metrics such as observed and Bayes error, and weakest for MAR, consistent with a loss of discriminative structure that hurts decisions without creating a systematic over- or under-confidence bias. The trefoil suggests a boundary of the complexity-driven regime: beyond a critical entanglement, added difficulty no longer enriches the representation but dismantles it.

\begin{table}[t]
\centering
\caption{Per-family Spearman correlation between $H_1$ area and observed error. Five families are positive; the trefoil knot inverts, indicating a collapse rather than a complexity-driven failure mode. ${}^{*}p<0.05$, ${}^{***}p<0.001$; n.s.\ not significant.}
\label{tab:perfamily}
\begin{tabular}{lc}
\toprule
Family & $\rho(H_1\text{ area},\ \text{obs.\ error})$ \\
\midrule
Hopf link & $0.83^{***}$ \\
Sphere & $0.61^{***}$ \\
Warped circle & $0.59^{***}$ \\
Swiss roll (control) & $0.43^{*}$ \\
Torus & $0.16$ (n.s.) \\
Trefoil knot & $-0.47^{**}$ \\
\bottomrule
\end{tabular}
\end{table}

\subsection{Behavior under extreme stress}
We also investigate an extreme suite at difficulty levels $6$--$8$, with noise up to $1.5$ and up to $200$ nuisance dimensions. Three observations support the moderate-difficulty interpretation. First, the trefoil's inverted relationship disappears: once the knot and its scaled copy overlap almost completely, the representation has already collapsed, leaving little residual topology--error relationship to measure. Second, the $H_0/H_1$ scissors persists. Across the extreme levels, $H_1$ area and the $H_0$ fragmentation count continue to rise with difficulty while $H_0$ total persistence continues to fall. Third, topology still predicts decision error strongly: $H_1$ area correlates with observed error at $\rho=0.84$. Its correlation with MAR, however, saturates toward zero, as expected once both the true probability and the model move toward chance under overwhelming noise. The extreme regime therefore separates two claims that are coupled at moderate difficulty: topology continues to track decision failure, while the calibration-residual signal saturates when the task itself approaches chance.

\subsection{Swiss roll as a negative control}
The Swiss roll is topologically equivalent to a two-dimensional sheet and has no true loop structure, despite being geometrically curved. It therefore serves as a negative $H_1$ control: if our $H_1$ descriptors were responding only to generic curvature, noise, or sampling artifacts, Swiss roll would show effects comparable to genuinely loop-bearing families such as the Hopf link or warped circle.

The Swiss roll does show some $H_1$ growth with increasing noise and a moderate topology--error correlation in Table~\ref{tab:perfamily}, but the effect is substantially weaker than for genuinely loop-bearing topologies such as the Hopf link. This is the expected leakage for a geometric control: noise can create spurious cycles, but the response is weaker than when the data contain true loop-like structure. The $H_0$ signal is similarly muted; its fragmentation count barely tracks MAR compared with the genuinely structured families. Both homological dimensions therefore respond more strongly to intrinsic topology than to curvature or noise alone.

\subsection[Timing of H1 resolution]{Timing of $H_1$ resolution}
\label{sec:timing}
\begin{figure}[t]
\centering
\begin{subfigure}{0.49\textwidth}\includegraphics[width=\linewidth]{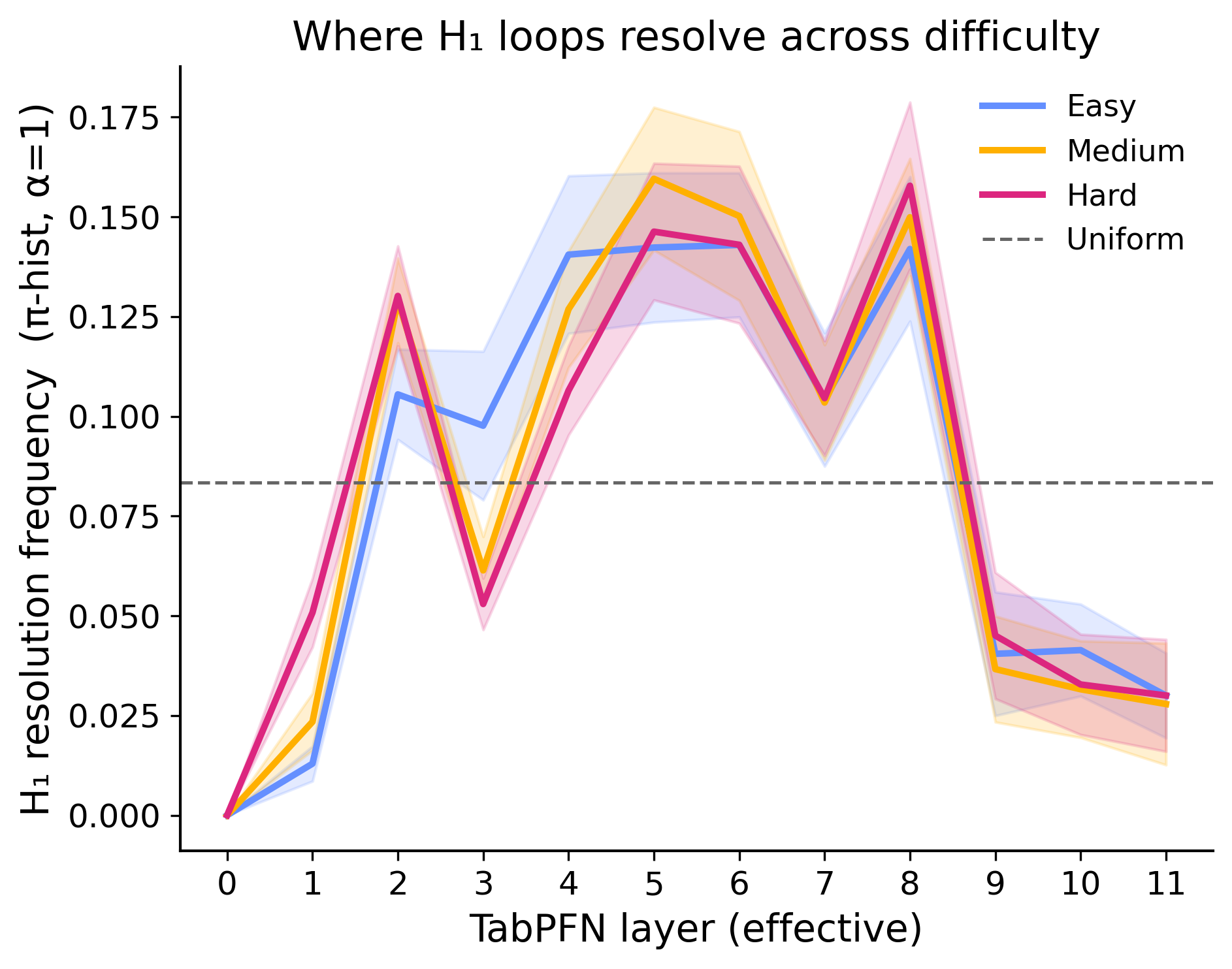}\caption{}\end{subfigure}\hfill
\begin{subfigure}{0.49\textwidth}\includegraphics[width=\linewidth]{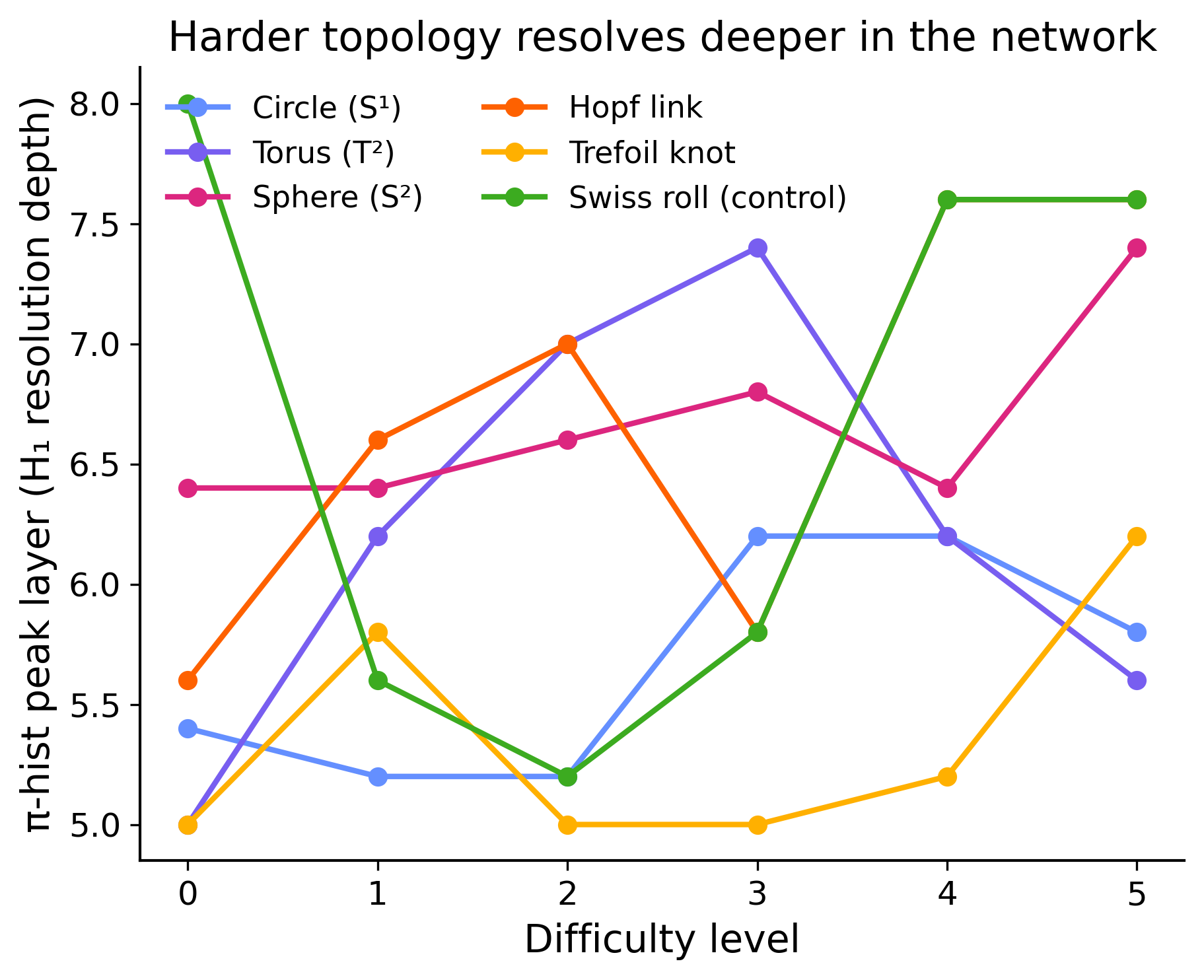}\caption{}\end{subfigure}
\caption{Layerwise timing of $H_1$ resolution. (a) Persistence-weighted birth histogram $\pi_{\mathrm{hist}}(\alpha=1)$ for easy, medium, and hard datasets. (b) The $\pi_{\mathrm{hist}}$ peak layer as a function of difficulty; entangled families resolve their loops deeper in the network. Each individual run's peak falls on one of the $12$ discrete TabPFN layers; the plotted curve is the mean peak layer over the five seeds, so it can take fractional values (e.g., a value of $6.5$ indicates that the peak lies between layers $6$ and $7$ when averaged across seeds).}

\label{fig:timing}
\end{figure}

\begin{figure}[t]
\centering
\begin{subfigure}{0.24\textwidth}\includegraphics[width=\linewidth]{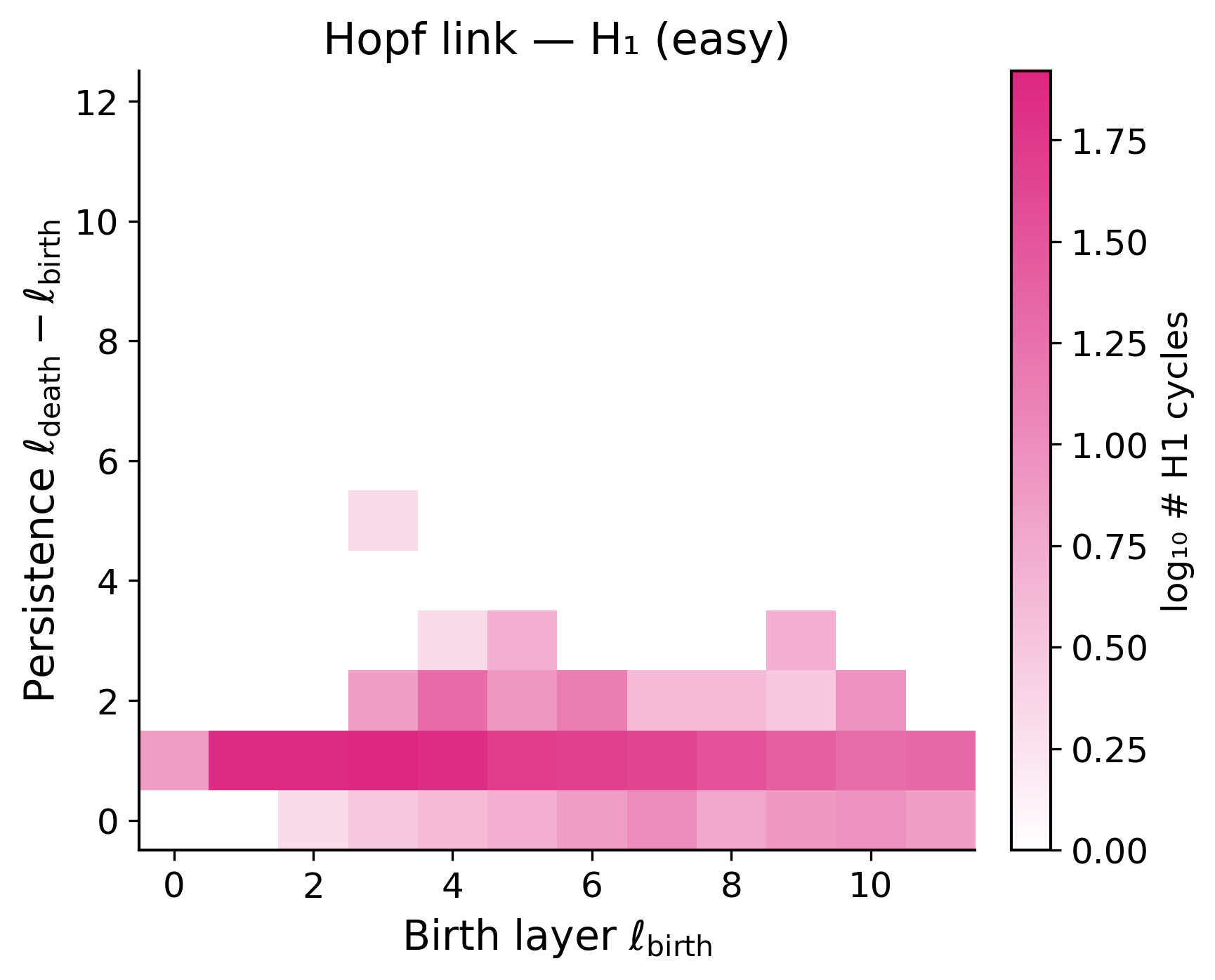}\caption{}\end{subfigure}\hfill
\begin{subfigure}{0.24\textwidth}\includegraphics[width=\linewidth]{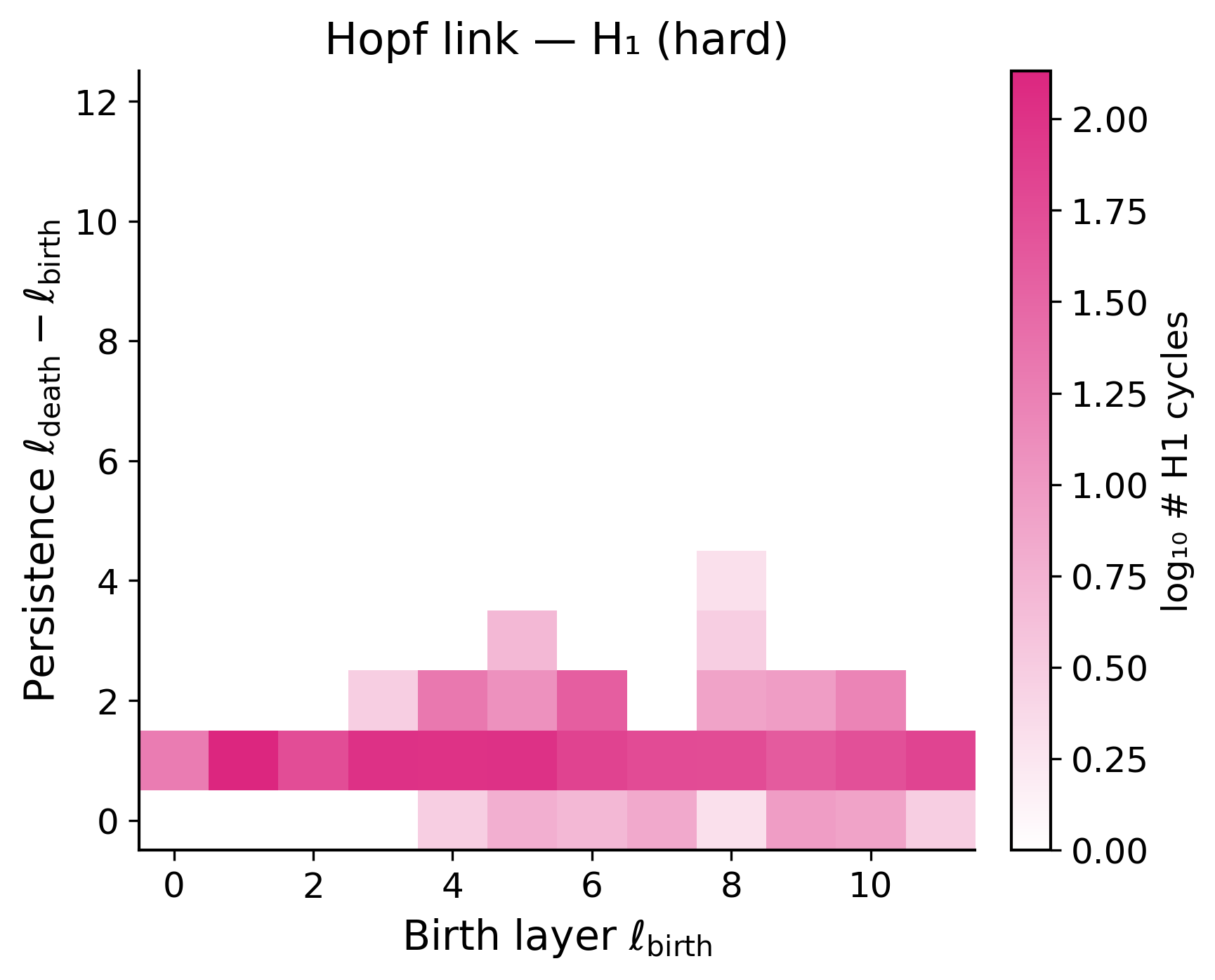}\caption{}\end{subfigure}\hfill
\begin{subfigure}{0.24\textwidth}\includegraphics[width=\linewidth]{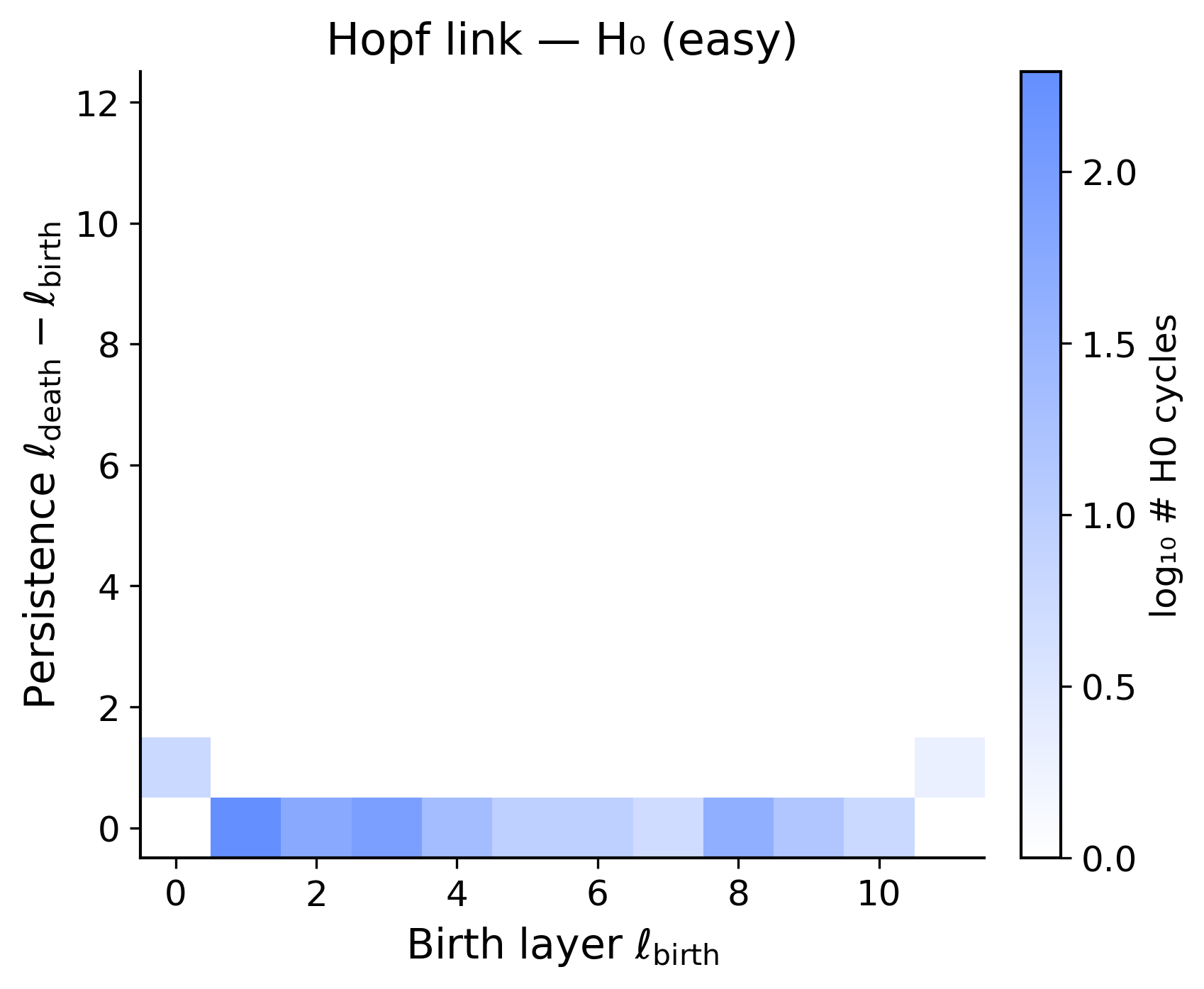}\caption{}\end{subfigure}\hfill
\begin{subfigure}{0.24\textwidth}\includegraphics[width=\linewidth]{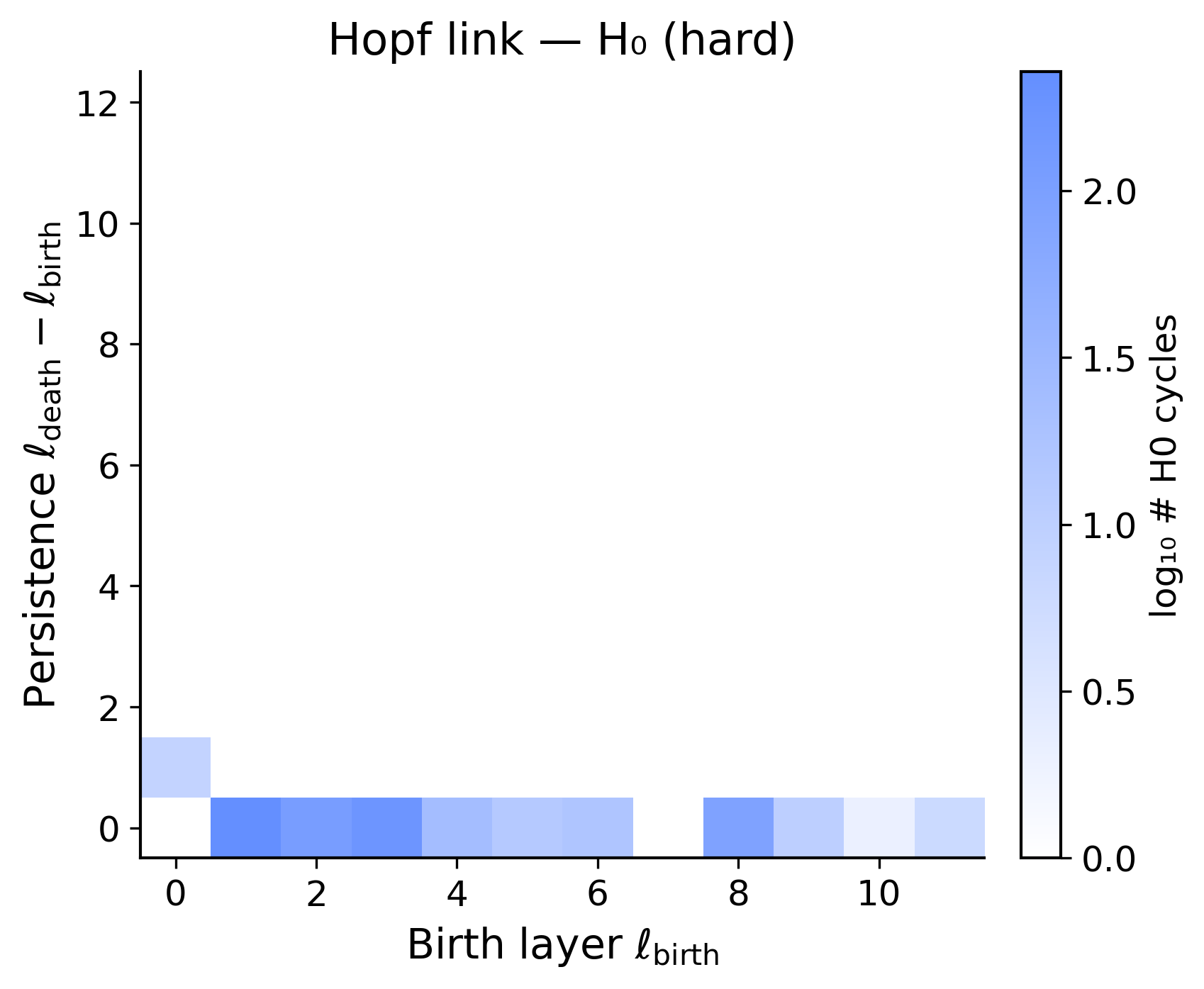}\caption{}\end{subfigure}
\caption{Birth--persistence heatmaps for the Hopf link (log count of cycles by birth layer and persistence). (a,b) $H_1$ for easy vs hard; (c,d) $H_0$ for easy vs hard. At the main benchmark scale, hard datasets show more persistent $H_1$ mass and a shift of $H_0$ mass toward short lifetimes.}
\label{fig:heatmaps}
\end{figure}

Using $\pi_{\mathrm{hist}}(\alpha=1)$ and birth-persistence heatmaps in Figure~\ref{fig:timing} and Figure~\ref{fig:heatmaps}, we analyze where $H_1$ features appear and resolve across TabPFN layers. For the entangled families, increasing difficulty shifts persistent $H_1$ activity toward later layers. The $\pi_{\mathrm{hist}}$ peak layer rises with difficulty for the trefoil knot and the Hopf link, while the simpler families show no significant shift. The depth shift is therefore specific to topologies that are genuinely hard to disentangle. At the main benchmark scale, the birth-persistence heatmaps corroborate this at the bar level: relative to easy datasets, hard datasets carry more persistent $H_1$ mass and shift their $H_0$ mass toward short lifetimes.

This qualitatively echoes prior LLM zigzag findings in \cite{gardinazzi2025persistent}, where long-lived $H_1$ features tend to appear in middle layers and late-layer changes reflect output-oriented rearrangement. The interpretation differs, however. In LLMs, topology describes how prompts are reorganized during language processing. In TabPFN, topology describes how an in-context tabular task geometry is constructed and whether that construction is reliable. Timing adds a depth axis to the same story: hard topology changes not only the amount and stability of topological structure, but also where in the network the surviving loop structure resolves.

\subsection{Fine-tuning under topological stress}
We also evaluate whether small-scale fine-tuning can repair hard topology regimes by comparing matched base and fine-tuned pairs \cite{tanna2026exploring}. No reliability metric improves significantly: pooled Wilcoxon signed-rank tests give $p=0.79$ for observed error, $p=0.56$ for Bayes error, and $p=0.62$ for MAR. Under extreme stress, the effect is slightly adverse. At the hardest difficulty level, fine-tuning raises the mean signed residual and the overconfidence rate, yielding a model that is more confident but no more correct.

Reduced reliability is therefore unlikely to be only a matter of insufficient task-specific adaptation. The limitation appears tied to the pretrained prior, or to the internal representation topology that difficult geometries induce.

\subsection{Robustness across embedding scopes}
Finally, we repeat the topology analysis in two additional embedding scopes: support feature-token representations and support label-token representations. The input-topology signal is stable across these scopes. The correlation between input noise and embedding $H_1$ count is $\rho=0.78$, $0.71$, and $0.72$ for the three scopes, respectively; for $H_0$ total persistence, the corresponding correlations are $\rho=0.83$, $0.82$, and $0.80$. The signal is therefore not an artifact of reading out one particular representation surface. Topological stress appears across the in-context task representation.

\section{Discussion}

\subsection{TabPFN topology as in-context task geometry}
The main conclusion is that TabPFN topology should be interpreted as the topology of an inferred task. TabPFN predicts by conditioning on the entire support-query context, and its internal representations encode that task-level geometry.

When input topology is simple and aligned with the model's learned prior, TabPFN can form stable coarse groupings. When topology becomes difficult, noisy, or poorly aligned with the prior, the internal topology becomes more fragmented and more cyclic. This produces the observed $H_0/H_1$ scissors effect. This view complements recent efforts to understand what TabPFN internalizes about a task, such as whether it recovers causal structure \cite{swelam2025does} or can be repurposed as a feature extractor \cite{ye2025closerlooktabpfnv2}; here the object of study is the reliability of the inferred task geometry rather than its predictive content.
This interpretation is consistent with recent evidence that tabular in-context generalization depends strongly on the prediction tasks induced during pretraining. The authors show that even a single real table can support broad transfer when it yields a sufficiently rich collection of self-supervised tasks \cite{ma2025generalization}. From this perspective, the geometry formed by TabPFN at inference time reflects an interaction between the supplied context and task-solving strategies encoded during pretraining. Our results suggest that topological fragmentation may arise when the new context induces a task geometry that is difficult to represent using those learned strategies.

\subsection{Comparison with other works}
Our work is methodologically close to the LLM zigzag persistence framework \cite{gardinazzi2025persistent}, which tracks evolving point clouds of hidden representations across transformer layers. That work identifies phases of prompt processing and uses topological descriptors for layer pruning.

In TabPFN, the same mathematical machinery reveals a different phenomenon. The most actionable signal is not layer redundancy, but dataset-level reliability. This difference is natural: TabPFN predictions are conditioned on an entire tabular context, so the topology of the inferred in-context task geometry becomes the relevant object of study.

Furthermore, HalluZig shows that topological signatures of evolving attention graphs can distinguish hallucinated from factual LLM generations \cite{samaga2026halluzig}. Our work differs in both object and setting. HalluZig analyzes attention topology in autoregressive language generation, while we analyze representation topology in tabular in-context inference. The common theme is that zigzag persistence can expose reliability-relevant structure in transformer internals, but the appropriate unit of analysis depends on the model and task.

\subsection{Practical implications}
The results suggest several applications and directions for future explorations. First, topology can serve as a dataset-level reliability diagnostic. Before trusting TabPFN on a new tabular context, one could compute internal topological descriptors and flag contexts that resemble high-risk regimes.

In addition, topology may be useful for monitoring population drift. If a deployed dataset gradually shifts toward more fragmented $H_0$ or more active $H_1$, that may indicate increasing reliability risk.

Lastly, the results motivate topology-aware pretraining or data augmentation. If TabPFN is less reliable on certain topological regimes, future pretraining distributions could explicitly include more complex or knotted tabular structures. 
Taken together with the findings of \cite{ma2025generalization}, our results raise the possibility that geometric and topological diversity constitutes one dimension of task richness. Increasing sample size may improve estimation within a given regime, but broader structural coverage may be required to improve generalization across regimes.
This perspective suggests a topology-aware approach to pretraining-task construction. Rather than increasing the number of synthetic rows while sampling from a fixed or limited family of generators, one could diversify the intrinsic geometry of the generated tasks, including connected, loop-bearing, linked, knotted, and geometrically curved but loop-free regimes. The objective would not be to reproduce particular manifolds literally, but to expand the range of representation structures that the model must organize during pretraining. Testing whether such coverage improves reliability on unseen complex geometries is an important direction for future work.

\subsection{Limitations and future work}
This study is primarily based on controlled synthetic data. This design is useful because the true conditional probabilities are known by construction, allowing us to measure MAR, Bayes error, and overconfidence relative to the data-generating process. Real-world tabular datasets may contain additional structure not captured by our generators, including heterogeneous feature types, missingness patterns, categorical variables, domain-specific constraints, distribution shifts, and non-geometric sources of label noise \cite{grinsztajn2022whytree,labelnoise_tabpfn}. The synthetic benchmark therefore provides a controlled stress test of TabPFN's behavior under known geometric and topological conditions, but it does not replace evaluation on diverse real-world benchmarks and tasks \cite{tabarena,tabdpt}.

A second limitation is that our topological summaries are coarse. The descriptors used in this study, such as $H_0$ and $H_1$ counts, total persistence, and birth-persistence summaries, characterize global properties of the representation topology. They are useful for identifying context-level reliability patterns, but they do not directly explain which support rows, query rows, or features are responsible for reduced reliability. In this sense, our analysis provides a diagnostic of the in-context task geometry as a whole rather than a mechanistic attribution method. Attention-based topology, especially query-to-support or feature-wise attention topology, may provide a more local view of how TabPFN routes information through the context \cite{samaga2026halluzig,hu2026noiseimmunityincontexttabular}.

The graph construction also introduces methodological choices. We follow prior zigzag representation work by building k-nearest-neighbor graphs at each layer, but the resulting topology depends on the choice of $k$, the distance metric, the embedding scope, and the maximum simplex dimension. Although our robustness checks suggest that the main context-level signal is not specific to one embedding scope, different graph constructions may emphasize different aspects of the representation geometry. In particular, k-nearest-neighbor graphs preserve relative neighborhood structure but do not explicitly encode absolute distances. Future work could compare k-nearest-neighbor complexes with distance-thresholded (e.g., Vietoris--Rips) constructions \cite{edelsbrunner2010computational,maria2014gudhi}.

Finally, TabPFN v2 \cite{hollmann2025tabpfnv2} has only 12 transformer layers, giving fewer depth positions than typical LLMs. This limits the resolution of layerwise phase analysis compared with recent zigzag studies of large language models, which often analyze 30 or more layers. The smaller depth makes broad early/middle/late trends visible, but it limits our ability to identify fine-grained processing phases. Future work could apply the same framework to deeper and larger tabular foundation models \cite{tabpfn_v2_5,tabpfn_v3,tabdpt,orion_msp}, compare multiple TabPFN versions, or analyze intermediate sub-blocks within each TabPFN layer, such as feature-wise attention, row-wise attention, and MLP updates, to obtain a finer view of the evolution of the in-context task geometry.

\section{Conclusion}
We introduced a zigzag persistent homology framework for analyzing TabPFN's internal representations across layers. Using controlled synthetic topologies with known true probabilities, we found that TabPFN's representation topology predicts dataset-level reliability. Harder input geometries induce a dual topological stress signature: increased $H_1$ loop activity and increased $H_0$ fragmentation, together with shorter-lived connected-component structure.
These descriptors correlate with Bayes error, MAR, and overconfidence. We further find that small-scale fine-tuning does not reliably repair these hard-topology regimes, indicating that the reduced reliability is tied to the pretrained prior rather than to insufficient task-specific adaptation.
Our results connect recent topological analyses of transformer representations to tabular foundation models and suggest a new direction for reliability diagnostics in TabPFN: topology-aware monitoring of the in-context task geometry.

\section*{Acknowledgments}
We would like to thank Travis Ens for his continued involvement throughout this project, from its early development to the revision of this manuscript, and for his valuable feedback and constructive suggestions, which helped improve the organization and presentation of this paper.

% --- References ---
\newpage
\bibliographystyle{plain}
\bibliography{refs}

% --- Appendix ---
\newpage
\appendix
This appendix provides additional details about the experimental setups used in the main paper.
\section{Experimental Details}
\subsection{Synthetic Data Generation}
\label{app:synthetic_generation}
All six topology families are produced by a common pipeline; they differ only in the base manifold, its embedding map, and the label score. For a fixed family, difficulty level, and random seed we (i)~sample points on a base manifold, (ii)~add isotropic off-manifold Gaussian noise, (iii)~append pure-noise nuisance dimensions, and (iv)~assign a smooth ground-truth label probability from the intrinsic manifold coordinates. Because the label is a known function of those coordinates, the true conditional probability $P^{*}(y\mid x)$ is available for every point, which is what makes the true-probability reliability metrics of the main text well defined.

Let $z$ denote the intrinsic coordinate(s) of a sampled point, $\Phi$ the family embedding into $\mathbb{R}^{2}$ or $\mathbb{R}^{3}$, and $\varsigma(u)=1/(1+e^{-u})$ the logistic function. Every dataset is constructed as
\begin{align}
x_{0} &= W\,\Phi(z) + \varepsilon, & \varepsilon &\sim \mathcal{N}(0,\sigma^{2} I), \\
x &= \big[\,x_{0}\,;\ \xi\,\big], & \xi &\sim \mathcal{N}(0, I_{m}), \\
\eta(z) &= \varsigma\!\big(\kappa\, g(z)\big), & y &\sim \mathrm{Bernoulli}\big(\eta(z)\big),
\end{align}
where $\sigma$ is the off-manifold noise level, $m$ the number of appended nuisance dimensions, $W$ an anisotropic warp that scales the first ambient axis by a factor $w$, $g(z)$ a family-specific score, and $\kappa$ the label sharpness. We fix $\kappa = 4.0$ in every run so that the intrinsic label (Bayes) noise is held constant and geometric difficulty is decoupled from label difficulty. For each point we store the true class probabilities $[\,1-\eta,\ \eta\,]$, the Bayes-optimal label $\mathbf{1}[\eta \ge \tfrac12]$, and the Bayes confidence $\max(\eta, 1-\eta)$. Unless otherwise noted, each dataset uses $n_{\mathrm{train}} = 400$ context rows and $n_{\mathrm{test}} = 300$ query rows, and every (family, level) pair is generated with five random seeds. The large-sample case study of the main text reuses the warped circle generator with $n_{\mathrm{train}} = 6667$ and $n_{\mathrm{test}} = 5000$ (and a half-scale $3333/2500$ variant), again with five seeds.

\paragraph{Family embeddings and label scores.}
Writing $t,\theta,\phi \sim \mathcal{U}(0,2\pi)$, $u\sim\mathcal{U}(-1,1)$, and $s\sim\mathcal{U}(0,1)$ for the intrinsic coordinates, the six families are:
\begin{itemize}
  \item \textbf{Warped circle} ($S^{1}\subset\mathbb{R}^{2}$): radius $r = 1 + \tau\sin(3\theta)$ and $\Phi(\theta) = \big(r\cos\theta,\ r\sin\theta\big)$; score $g=\sin\theta$. The twist $\tau$ deforms the radius at frequency $3$ while preserving the homotopy type ($S^{1}$).
  \item \textbf{Torus} ($T^{2}\subset\mathbb{R}^{3}$): with $R=3$ and $r=1$, $\Phi(\theta,\phi) = \big((R + r\cos\theta)\cos\phi,\ (R + r\cos\theta)\sin\phi,\ r\sin\theta\big)$; score $g=\sin\phi$.
  \item \textbf{Sphere} ($S^{2}\subset\mathbb{R}^{3}$): with $\rho=\sqrt{1-u^{2}}$, $\Phi(u,v) = \big(\rho\cos v,\ \rho\sin v,\ u\big)$ sampled uniformly on the sphere; score $g=u$, a smooth hemisphere split by latitude.
  \item \textbf{Hopf link} (two linked $S^{1}\subset\mathbb{R}^{3}$): half of the points lie on ring $A=(\cos t,\ \sin t,\ 0)$ and half on ring $B=(\delta,\ \cos t,\ \sin t)$, where $\delta$ is the inter-ring separation; the score $g = \lVert x-c_{B}\rVert - \lVert x-c_{A}\rVert$ contrasts distances to the ring centers $c_{A}=(0,0,0)$ and $c_{B}=(\delta,0,0)$, so the label encodes ring membership.
  \item \textbf{Trefoil knot} (knotted $S^{1}\subset\mathbb{R}^{3}$): the base knot is $\gamma(t)=\big(\sin t + 2\sin 2t,\ \cos t - 2\cos 2t,\ -\sin 3t\big)$; class~$0$ is $\gamma(t)$ and class~$1$ is the scaled, shifted copy $s_{\mathrm{f}}\,\gamma(t)+0.2$. The score $g = \lVert x_{1:3}\rVert - m$ thresholds the ambient radius at the per-dataset median $m$ of the training radii, which keeps the two classes balanced at every difficulty level; a fixed threshold instead skews toward one class as the copies overlap at high difficulty.
  \item \textbf{Swiss roll} ($\mathbb{R}^{2}$ sheet $\subset\mathbb{R}^{3}$, negative control): with angle $a = 2\pi\,n_{\mathrm{turns}}\,s + \pi$ and independent height $h\sim\mathcal{U}(0,10)$, $\Phi(s,h) = \big(a\cos a,\ h,\ a\sin a\big)$; score $g = s - \tfrac12$. The sheet has no intrinsic loop, so any $H_{1}$ signal is noise-induced.
\end{itemize}
In every family, the first ambient axis of the embedding is multiplied by the warp factor $w$ before noise is added.

\paragraph{Difficulty schedule.}
Each family is escalated across nine ordered levels: levels $0$--$5$ form the main suite with $6\times6\times5 = 180$ runs and levels $6$--$8$ the extreme suite with $6\times3\times5 = 90$ runs. All families share the noise, nuisance, and warp ladder of Table~\ref{tab:app_levels}, with two minor exceptions: the sphere uses $\sigma=0.30$ at level~$4$, and the torus uses $\sigma=0.30$ at level~$4$ and $\sigma=0.50$ at level~$5$. Beyond this shared ladder, each structured family also tightens one geometry-specific knob that drives its components together or increases its curvature; the torus and sphere have no secondary knob and are stressed by noise, nuisance, and warp alone.

\begin{table}[t]
\centering
\caption{Shared difficulty ladder used by all six families. Levels $0$--$5$ are the main suite; levels $6$--$8$ are the extreme suite. The sphere uses $\sigma=0.30$ at level~$4$, and the torus uses $\sigma=0.30$ at level~$4$ and $\sigma=0.50$ at level~$5$.}
\label{tab:app_levels}
\begin{tabular}{lccccccccc}
\toprule
Level & 0 & 1 & 2 & 3 & 4 & 5 & 6 & 7 & 8 \\
\midrule
Noise $\sigma$    & 0.02 & 0.05 & 0.10 & 0.18 & 0.28 & 0.45 & 0.70 & 1.00 & 1.50 \\
Nuisance dims $m$ & 0 & 0 & 4 & 10 & 20 & 30 & 50 & 100 & 200 \\
Warp $w$          & 1.0 & 1.0 & 1.2 & 1.5 & 1.8 & 2.0 & 2.5 & 3.0 & 3.5 \\
\bottomrule
\end{tabular}
\end{table}

\begin{table}[t]
\centering
\caption{Family-specific secondary difficulty knob by level. Smaller ring separation and smaller trefoil scale factor push the two components together, while more Swiss-roll turns increase curvature. The torus and sphere have no secondary knob.}
\label{tab:app_knobs}
\begin{tabular}{lccccccccc}
\toprule
Level & 0 & 1 & 2 & 3 & 4 & 5 & 6 & 7 & 8 \\
\midrule
Warped circle: twist $\tau$            & 0.00 & 0.10 & 0.20 & 0.30 & 0.45 & 0.60 & 0.80 & 1.00 & 1.20 \\
Hopf link: separation $\delta$         & 1.5 & 1.2 & 1.0 & 0.8 & 0.5 & 0.3 & 0.2 & 0.1 & 0.05 \\
Trefoil: scale factor $s_{\mathrm{f}}$ & 2.0 & 1.8 & 1.5 & 1.3 & 1.15 & 1.05 & 1.02 & 1.01 & 1.005 \\
Swiss roll: turns $n_{\mathrm{turns}}$ & 1.0 & 1.5 & 2.0 & 2.5 & 3.5 & 5.0 & 7.0 & 10.0 & 15.0 \\
\bottomrule
\end{tabular}
\end{table}

\subsection{Model configuration.} Newer TabPFN releases include v2.5 and v3 models \cite{tabpfn_v2_5, tabpfn_v3}. In this study we use a TabPFN v2 checkpoint with feature group size $=1$, similar to the setup in \cite{ye2025closerlooktabpfnv2, hu2026noiseimmunityincontexttabular} to remove artifacts from estimator ensembling. To reduce randomness from inference-time ensembling, we use one estimator and disable feature shuffling when fitting the model.

\subsection{Compute resources.} All computational tests reported in this paper were conducted using a single NVIDIA Tesla T4 GPU.

\end{document}